\pdfoutput=1
\documentclass[10pt,twocolumn,letterpaper]{article}

\usepackage{geometry}
\usepackage{cite}
\usepackage{graphicx}
\usepackage{multirow}
\usepackage{url}
\usepackage{amsmath}
\usepackage{amsfonts}
\usepackage{bm}
\usepackage{xcolor}

\usepackage[ruled,vlined]{algorithm2e}

\SetCommentSty{mycommfont}
\SetKwRepeat{Struct}{struct \{}{\}}

\usepackage{fancyhdr}
\fancypagestyle{main}{%
   \fancyhf{} 
   
   \lhead{Submitted to IEEE Trans. Image Process.}
   \rhead{\thepage}
}%

\usepackage[percent]{overpic}
\usepackage[outline]{contour}
\usepackage{enumitem}
\usepackage{array}

\begin{document}

\title{\textbf{Parallel Discrete Convolutions on Adaptive Particle Representations of Images}}
\author{Joel Jonsson$^{1,2,3,5}$, Bevan L. Cheeseman$^{1,2,3,\dag}$, Suryanarayana Maddu$^{1,2,3,5}$, Krzysztof Gonciarz$^{2,3}$, \\ Ivo F. Sbalzarini$^{1,2,3,4,5,\ast}$ \\
	\normalsize{$^{1}$ Technische Universit\"{a}t Dresden, Faculty of Computer Science, 01069 Dresden, Germany} \\
	\normalsize{$^{2}$ Max Planck Institute of Molecular Cell Biology and Genetics, 01307 Dresden, Germany}\\
	\normalsize{$^{3}$ Center for Systems Biology Dresden, 01307 Dresden, Germany}\\
	\normalsize{$^{4}$ Cluster of Excellence Physics of Life, Technische Universit\"{a}t Dresden, Germany}\\
	\normalsize{$^{5}$ Center for Scalable Data Aanlytics and Artificial Intelligence ScaDS.AI, Dresden/Leipzig, Germany}\\
	\normalsize{$^\dag$Now at: ONI, Inc., Linacre House, Banbury Road, Oxford, OX2 8TA, UK.} \\
	\normalsize{$^\ast$Correspondence to: sbalzarini@mpi-cbg.de}
}

\date{}
\maketitle

\pagestyle{main}

\begin{abstract}
We present data structures and algorithms for native implementations of discrete convolution operators over Adaptive Particle Representations (APR) of images on parallel computer architectures. The APR is a content-adaptive image representation that locally adapts the sampling resolution to the image signal. It has been developed as an alternative to pixel representations for large, sparse images as they typically occur in fluorescence microscopy. It has been shown to  reduce the memory and runtime costs of storing, visualizing, and processing such images. This, however, requires that image processing natively operates on APRs, without intermediately reverting to pixels. Designing efficient and scalable APR-native image processing primitives, however, is complicated by the APR's irregular memory structure. Here, we provide the algorithmic building blocks required to efficiently and natively process APR images using a wide range of algorithms that can be formulated in terms of discrete convolutions. We show that APR convolution naturally leads to scale-adaptive algorithms that efficiently parallelize on multi-core CPU and GPU architectures. We quantify the speedups in comparison to pixel-based algorithms and convolutions on evenly sampled data. We achieve pixel-equivalent throughputs of up to 1\,TB/s on a single Nvidia GeForce RTX 2080 gaming GPU, requiring up to two orders of magnitude less memory than a pixel-based implementation. 
\end{abstract}

\section{Introduction}

Fluorescence microscopy enables long-term imaging of biological specimen at high spatial and temporal resolution with morphologically or biochemically specific labeling. This enables researchers to study biological structures and processes in living organisms~\cite{reynaud2014guide,mcdole2018toto,huisken2012slicing}, providing an invaluable source of information. However, handling and analyzing the large amounts of data produced causes significant computational demands, especially for three-dimensional (3D) images. Light-sheet microscopes, e.g., can acquire data at rates on the order of 10\,TB per hour~\cite{chhetri2015whole}, challenging image storage~\cite{allan2012omero}, visualization \cite{gunther2019scenery}, and processing~\cite{schindelin2012fiji}. This ``data bottleneck'' often limits the throughput and scalability of fluorescence microscopy studies and leads to under-utilization of the information contained in the images. 

To relax the data bottleneck, mainly three approaches are followed: (1) parallelization of image processing, (2) multi-resolution image representations, and (3) data compression. Parallelizing large images in distributed-memory computer clusters has enabled real-time segmentation of fluorescence microscopy images \cite{afshar2016parallel}, and GPU acceleration has enabled interactive handling of large images \cite{royer2015clearvolume, haase2020clij}. Multi-resolution representations and chunked hierarchical file formats aid analysis by allowing zoomable navigation of large volumes \cite{moore2021ome}. Tools like BigDataViewer \cite{pietzsch2015bigdataviewer} and TeraFly \cite{bria2016terafly} have leveraged this to enable interactive visualization and annotation of Terabyte-sized volumetric images. Compression methods, such as the Keller Lab Block (KLB) \cite{amat2015efficient} and B\textsuperscript{3}D \cite{balazs2017real} formats, reduce the cost of storing and transferring data. However, compression does not reduce the runtime of image processing pipelines, as the data must be decompressed to its original size prior to processing. 

Adaptive-resolution image representations, such as the Adaptive Particle Representation (APR) \cite{cheeseman2018adaptive}, provide an alternative to compression that can simultaneously reduce both data size and processing times. In order to take full advantage of such representations, however, it is necessary to natively process the images in the data-reduced representation, ideally on parallel computer architectures. This enables efficient end-to-end pipelines that never have to revert to pixels, not even block-wise. For certain tasks, like graph-cut segmentation, this has been shown to reduce both memory consumption and runtime \cite{cheeseman2018adaptive}. However, biological image analysis workflows typically require a wider range of algorithms. Adapting these algorithms, which have been designed for uniform pixel grids, to content-adaptive image representations, and efficiently implementing them using sparse data structures on parallel computers, is not trivial. Particularly challenging is the efficient implementation on GPUs, which are highly optimized for processing uniform grids.

Here, we provide the algorithmic building blocks for natively processing adaptive-resolution APR images in parallel on both CPUs and GPUs. Most classic low-level vision algorithms can be formulated as (sequences of) discrete convolutions. We therefore propose algorithmic strategies for efficient implementation of discrete convolution filters on APRs. We further show how the varying spatial scales of the APR can be exploited to naturally define scale-adaptive filters. Benchmark results demonstrate the efficiency of our implementations compared to pixel-based filtering, and we present an example of image deconvolution using the Richardson-Lucy algorithm natively implemented on the APR.

\section{Background}

For convenience, we recapitulate the fundamental concepts of the APR, but refer to Ref.~\cite{cheeseman2018adaptive} and its Supplementary Material for more details and for the algorithms required to construct APRs from pixel images.

\subsection{The Adaptive Particle Representation}

\begin{figure}[h]
	\centering
    \begin{tabular}{c c}
        \textbf{Pixel image} & \textbf{APR} \\
        \includegraphics[width=0.45\linewidth]{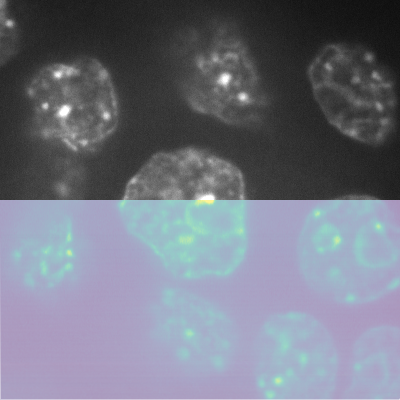} &
        \includegraphics[width=0.45\linewidth]{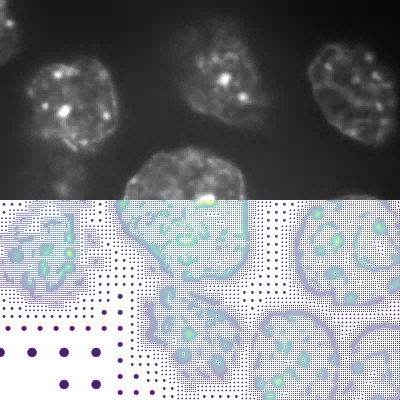}
    \end{tabular}
    
    \caption{Illustration of a fluorescence microscopy pixel image and a corresponding Adaptive Particle Representation (APR). The image used is a crop of Hoechst-stained mouse blastocyst cells \cite{rivron2018blastocyst}, available as image set BBBC032v1 from the Broad Bioimage Benchmark Collection \cite{ljosa2012annotated}. The top half of each panel shows the original image (left) and the image reconstructed from the APR (right). The bottom halves show the sampling points (pixels or particles) as dots, colored by sample value and scaled according to their spatial extent. Pixels are  uniform across the domain, whereas the APR particles adapt to the content of the image.}
	\label{fig:aprvspix}
\end{figure}

The APR optimally adapts the local density of the sampling $\mathcal{P}=\{(\mathbf{x}_p, f(\mathbf{x}_p))\}_{p=1}^{N_p}$ of some function, signal, or image, $f:\mathbb{R}^n \rightarrow \mathbb{R}$, defined over $\Omega \subset \mathbb{R}^n$ so as to guarantee that the point-wise reconstruction error, for all of a wide-class of reconstruction methods $\hat{f}$, is bounded everywhere in $\Omega$ by a user-set constant $E$ relative to a (potentially spatially varying) local error scale. This places sampling points $\mathbf{x}_p$ where they are required in order to approximate the unknown continuous $f$ within the given error threshold at all original pixel locations. Figure~\ref{fig:aprvspix} illustrates the result for a fluorescence microscopy image. In regions where the signal gradient is significant, the original (pixel) sampling density is retained. However, in regions of low signal gradient, such as the background or homogeneous areas in the interior of objects, the sampling density is reduced.

Mathematically, the APR bounds the infinity norm of the relative reconstruction error
\begin{align}
	\left\|\frac{f - \hat{f}}{\sigma}\right\|_{\infty} \leq E , \label{eq:r_con}
\end{align}
where $f$ is the unknown true signal, $\hat{f}$ the reconstruction, and $\sigma$ the local error scale. The reconstruction $\hat{f}$ at location $\mathbf{x}$ can be any positive weighted combination of sampled function values $f(\mathbf{x}_p)$ within a certain radius $R(\mathbf{x})$ of $\mathbf{x}$, where the function $R : \Omega \rightarrow \mathbb{R}$ is called {\em resolution function}. Thus, for an APR, at any location $\mathbf{x}\in\Omega$, any reconstruction of the form
\begin{align}
	\hat{f}(\mathbf{x}) = \sum_{\mathbf{x}_p : |\mathbf{x} - \mathbf{x}_p| \leq R(\mathbf{x})}f(\mathbf{x}_p)w_p, 
	\label{eq:interp}
\end{align}
with $\sum_p w_p = 1$ and $w_p \geq 0$\footnote{The non-negativity constraint can be relaxed, and classes of adaptation satisfying higher-order constraints can also be formulated \cite{cheeseman2018adaptive}} fulfills the reconstruction condition in Eq.~\ref{eq:r_con}. The resolution function can intuitively be seen as a local length scale of the signal.

If we consider a signal originally evenly sampled at $N$ points then, in general dimensions and for arbitrary $R(\mathbf{x})$ and $\mathcal{P}$, only greedy locally optimal solutions can be found in quadratic runtime $O(N^2)$, becoming infeasible even for small problems \cite{cheeseman2018adaptive}. In the APR both the resolution function $R(\mathbf{x})$ and the sampling locations $\mathbf{x}_p$ are therefore restricted to be power-of-two fractions of the image edge length $|\Omega|$. Under these restrictions, globally optimal sampling solutions can be found in linear time $O(N)$ \cite{cheeseman2018adaptive}. The use of power-of-two decompositions is common in image representations and image processing algorithms, such as image pyramids \cite{adelson1984pyramid}, tree-based methods \cite{meagher1982geometric}, and wavelet decompositions \cite{porwik2004haar}.

\begin{figure}[h!]
    \begin{tabular}{l l}
         \textbf{A} \hfill
         &
         \textbf{B} \hfill
         \\
         \centering
         \includegraphics[width=.45\linewidth]{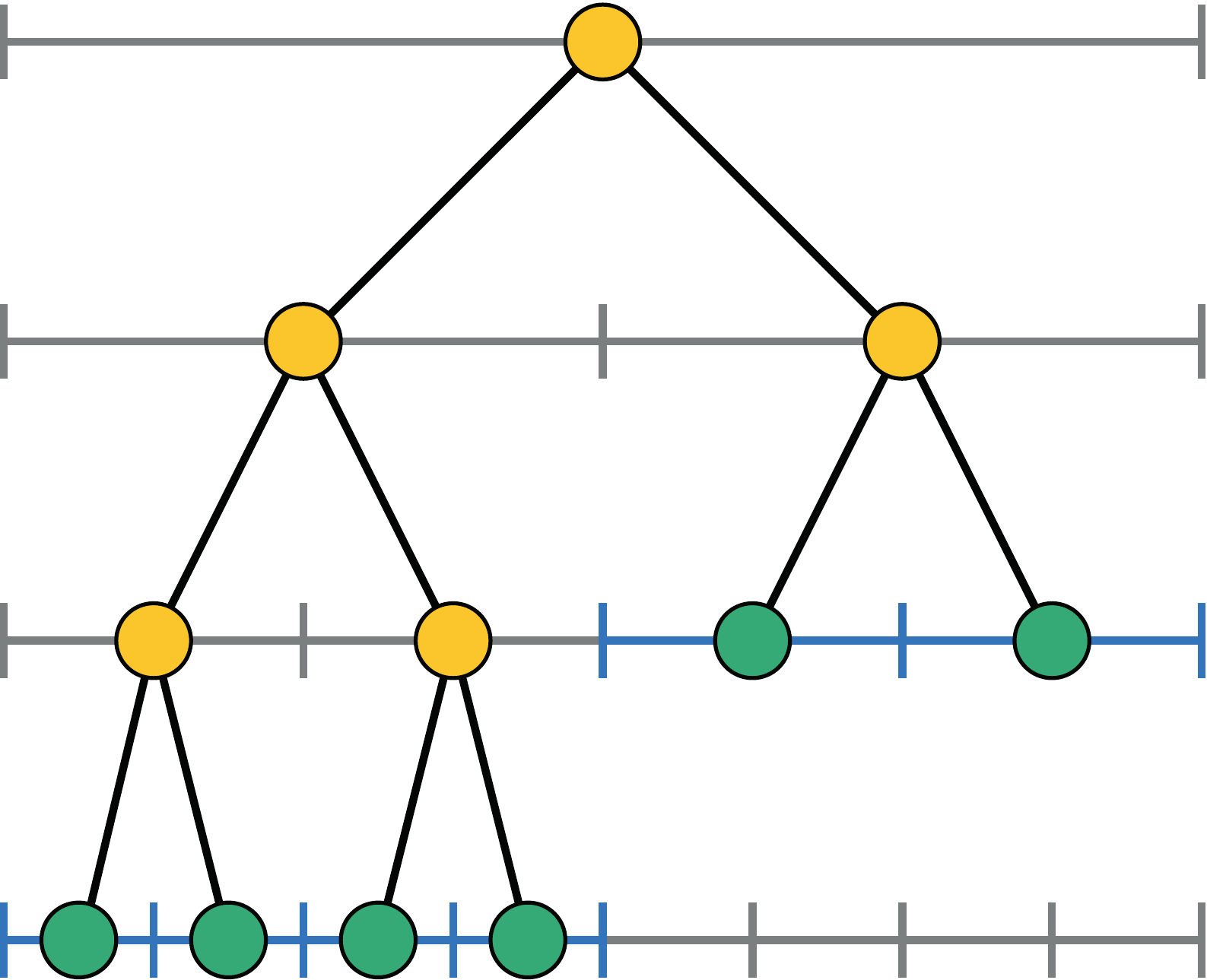}
         &
         \centering
         \includegraphics[width=.45\linewidth]{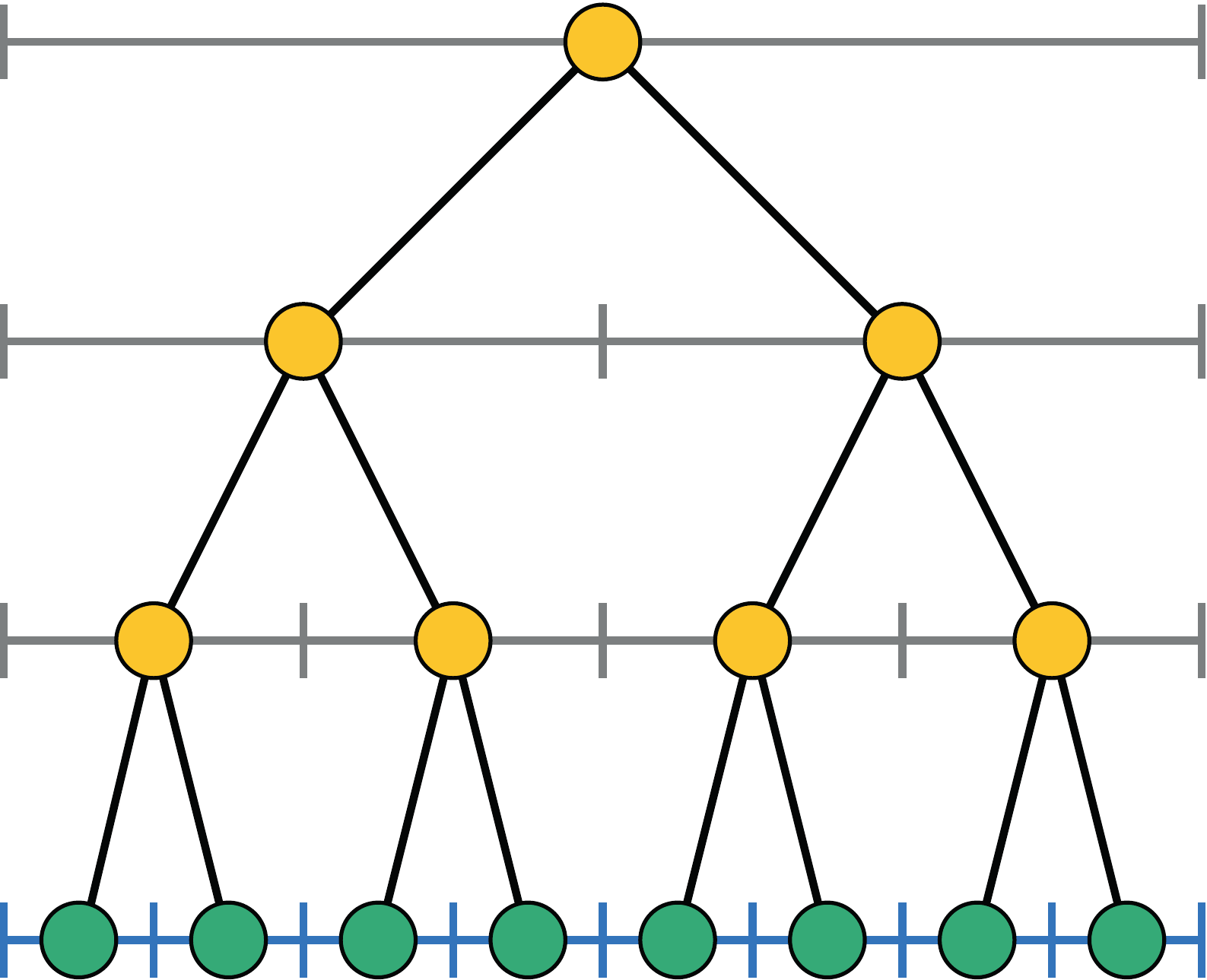}
    \end{tabular}
	\caption{Schematic comparing a regular sampling (pixels) to an APR in one dimension. (A) shows the pixels as green dots, and successively downsampled (by factors of 2) super-pixels as yellow dots. This forms an image pyramid or, by connecting spatially overlapping elements, a full binary tree. The APR particles (green dots in B) define a partition of the image domain and coincide with nodes of this tree at different levels of resolution.}
	\label{fig:compare}
\end{figure}

Considering the analogy of an image pyramid for a function in one dimension, as illustrated in Fig.~\ref{fig:compare}, the APR corresponds to a partition of the domain $\Omega$ across resolution levels $l$, where at each location the coarsest element is selected under the condition that Eq.~\ref{eq:r_con} holds. Intuitively, this can be thought of as a pruning operation of the full tree, where branches in areas of low signal gradient (relative to $\sigma$) are cut, and the sampling values are pushed to coarser ascendant nodes. The APR particles (green dots in Fig.~\ref{fig:compare}B) thus constitute the leaf nodes in a pruned tree structure. The locations of particles are taken to be the centers of the corresponding grid cells (blue intervals in Fig.~\ref{fig:compare}B), which we refer to as {\em particle cells}. The particle cells partition the image domain, and correspond exactly to grid cells in the image pyramid. In this way, they can be described by their resolution level $l_p$ and a multi-index $\mathbf{i}_p=(i_{p,1},\ldots ,i_{p,n})$ defining the location of the cell in the grid at level $l_p$. This allows the location $\mathbf{x}_p$ of a particle cell to be written in terms of $\mathbf{i}_p$ and $l_p$ as
\begin{align}
	\mathbf{x}_p = \mathbf{i}_p\frac{|\Omega|}{2^{l_p}}.
	\label{eq:part_loc}
\end{align}
For simplicity we assume the image domain $\Omega$ to be a hypercube starting at $\mathbf{0}$ with all edges of length $|\Omega|$ a power of two. For anisotropic domains, or domains that are not a power of two, $|\Omega|$ is the maximum edge length extended accordingly.

\subsection{Neighborhood and Reconstruction}
\label{sec:recon}

The resolution function $R(\mathbf{x})$ is taken to be a piecewise constant function, defined in terms of the APR particles as:
\begin{equation}
    \label{eq:resolution_function}
	R(\mathbf{x}) = \sum_{p=1}^{N_p}\frac{|\Omega|}{2^{l_p}}\left[\left\lfloor\frac{\mathbf{x}}{2^{l_p}}\right\rfloor=\mathbf{i}_p\right],
\end{equation}
where $[A] = 1$ if $A$ is true and $0$ otherwise. Thus, the value of the resolution function $R(\mathbf{x})$ at any location $\mathbf{x} \in \Omega$ is determined solely by the level of the particle cell which contains $\mathbf{x}$. This is guaranteed to be unique, since the particle cells partition the image domain $\Omega$. In the example of Fig.~\ref{fig:compare}B, evaluating $R(x)$ corresponds to finding the level $l$ for which the blue interval containing $x$ holds a particle (green dot). The resolution function then takes the value $R(x)=\frac{|\Omega|}{2^{l}}$, which defines the maximum radius of the reconstruction neighborhood in Eq.~\ref{eq:interp}. 

The reconstruction condition in Eq.~\ref{eq:r_con} is valid for any positive weighted combination of particles within the radius $R(\mathbf{x})$. This affords a lot of flexibility in defining different reconstruction methods. However, in image processing algorithms we have found that it is often favorable to use the simplest possible method. That is, to reconstruct the signal at a location $\mathbf{x}$, we find the particle $(\mathbf{x}_p^*, f(\mathbf{x}_p^*))$ whose particle cell contains $\mathbf{x}$ and take $\hat{f}(\mathbf{x}) = f(\mathbf{x}_p^*)$. We refer to this as piecewise constant reconstruction. Compared to more general reconstructions, this method is computationally efficient to evaluate, as the reconstructed values do not depend on neighbouring particles. Moreover, this allows an intuitive view of particles and particle cells as pixels of different sizes. In the remainder of this text, unless otherwise stated, all reconstructions are assumed to be piecewise constant.

\subsection{Determining the APR}

Computing the APR from a pixel image requires estimating the gradient magnitude of the intensity field, $|\nabla f|$, as well as the local error scale $\sigma(\mathbf{x})$ at the original $N$ pixels. These are then combined to form
\begin{align}
	L(\mathbf{x})=\frac{E\sigma(\mathbf{x})}{|\nabla f|},
\end{align}
which is quantized into a tree structure, and the APR solution is computed using a linear-time algorithm called the {\em Pulling Scheme} \cite{cheeseman2018adaptive}. This algorithm directly provides the adaptive tree structure as shown in Fig.~\ref{fig:compare}B, i.e., the set of particle cells, which implicitly define both the resolution function $R(\mathbf{x})$ and the particle locations $\{\mathbf{x}_p\}_{p=1}^{N_p}$ of $\mathcal{P}$. The final step requires determining the particle intensity values $f(\mathbf{x}_p)$. Since particles at the finest resolution coincide with the original pixels, those values remain unchanged. Intensities at coarser particle locations are resampled. This can be done in a number of ways. Here, as well as in Ref.~\cite{cheeseman2018adaptive}, coarse particle values are determined by average downsampling the pixel values. This is simple and provides an inhered denoising effect. Viewed as an operation on the tree structure, coarse particle values (i.e., leaf nodes of pruned branches in the tree) are thus obtained by propagating the values of original leaf nodes (i.e., pixels) upward, level by level, taking the average of the combined nodes at each step.  

\section{Data structures and algorithms}

We start by detailing data structures and algorithmic strategies that can be used to define and efficiently implement a wide range of image processing algorithms to natively work on the APR. This requires sparse tree data structures with the corresponding iterators, as well as local isotropic patch reconstruction. 

\subsection{Sparse APR data structure}

As described in Eq.~\ref{eq:part_loc}, the location $\mathbf{x}_p$ of a particle (cell) is defined by its level $l_p$ and multi-index $\mathbf{i}_p$ encoding the cell coordinates in the grid at the corresponding resolution level. This definition offers a lot of flexibility in choosing data structures for storing the APR. Here, we base our design on the analogy with image pyramids, as illustrated in Fig.~\ref{fig:compare}B. Since the particle cells partition the image domain, the APR corresponds to a pyramid of disjoint sparse images, where each location $\mathbf{x} \in \Omega$ is covered by exactly one particle cell at some resolution level. From this perspective, the APR amounts to a set of sparse images at different resolutions, which can be encoded using any sparse array format.

The original APR data structures~\cite{cheeseman2018adaptive} store the particle values $\{f(\mathbf{x}_p)\}_{p=1}^{N_p}$ as a single, contiguous vector, while the spatial coordinates $\{\mathbf{x}_p\}_{p=1}^{N_p}$ are encoded as follows: The levels and all but one of the spatial dimensions are stored in a dense array with each ``row'' in the contiguous dimension\footnote{In order to stay consistent with the software libraries, we  assume that the image dimensions are ordered as $(z, x, y)$, where $y$ is contiguous in memory, i.e., that values at locations $(z, x, y)$ and $(z, x, y+1)$ are adjacent in memory.} compressed. The sparse compression is done by storing the first and last index of each contiguous block of particles, along with a pointer to the particle values, in a red-black tree structure. This allows for efficient random access via red-black tree search in the sparse dimension.

While efficient random access is important in some applications, we argue that it is not required for most image processing tasks. Instead, these typically rely on the ability to iterate over neighborhoods of certain, fixed structure. Therefore, we here introduce a simplified data structure that explicitly stores the $y$-coordinates. This is equivalent to storing each resolution level $l$ in compressed sparse row (CSR) format and concatenating the vectors of row offsets and $y$-indices for the different levels. Algorithm~\ref{alg:linear_access} outlines the resulting data structure used to encode the spatial coordinates of particles. The particles are indexed linearly in the order $l \rightarrow z \rightarrow x \rightarrow y$, with sparse compression along $y$. Coordinate indices in the sparse dimension are stored in the vector {\tt y\_idx}, while the vector {\tt xz\_end} encodes the last particle index in each sparse row, with one entry for each combination of $(l,z,x)$. An additional vector {\tt level\_offset} stores the starting point of each level in the {\tt xz\_end} vector. The metadata required to correctly access these vectors are the minimum and maximum resolution levels, as well as the grid dimensions at each level.

\begin{algorithm}[h!]
    \label{alg:linear_access}
    \DontPrintSemicolon
    \KwSty{class} LinearAccess: \;
    \tcc{minimum and maximum resolution levels}
    $l_\mathrm{min}$, $l_\mathrm{max}$\;
    \tcc{image dimensions for each level}
    {\tt z\_dim[]}, {\tt x\_dim[]}, {\tt y\_dim[]}\;
    \tcc{sparse access vectors}
    {\tt y\_idx[]} \tcp{y index of each particle}
    {\tt xz\_end[]} \tcp{last particle of each row}
    {\tt level\_offset[]} \tcp{offsets in xz\_end per $l$} 
    
    \tcc{helper function to access a given row}
    \SetKwProg{Fn}{Function}{:}{}
    \Fn{get\_row(l, z, x)}{
        row $\gets$ {\tt level\_offset[}$l${\tt ]} + $z$*{\tt x\_dim[}$l${\tt ]} + $x$\;
        row\_begin $\gets$ {\tt xz\_end[}row$-1${\tt ]}\;
        row\_end $\gets$ {\tt xz\_end[}row{\tt ]}\;
        
        \KwRet row\_begin, row\_end\;
    }
    \BlankLine
    \caption{Linear APR access data structure}
\end{algorithm}

This linear access data structure complements, rather than replaces, the random access data structure from Ref.~\cite{cheeseman2018adaptive}. Both of the data structures are available in LibAPR \cite{libapr} (see~\ref{sec:code}), and the similarity between them allows for efficient conversion from one to the other depending on the anticipated access pattern. 

\subsection{Sequential data access and iteration}

Using the linear access data structure from Algorithm \ref{alg:linear_access}, querying particles at random locations is inefficient, but iterating sequentially over particles in the same sparse row is efficient, since both $y$-indices and particle properties are read from contiguous memory, promoting cache efficiency. Algorithm~\ref{alg:linear_iteration} illustrates how one would iterate over all particles of a given APR of class {\tt LinearAccess}, accessing both spatial coordinates and intensity values.

\begin{algorithm}[h!]
    \DontPrintSemicolon
    \KwData{LinearAccess {\tt apr}, {\tt intensity[]}}
    
    \For{($l$={\tt apr}.$l_\mathrm{min}$; $l<=${\tt apr}.$l_\mathrm{max}$; $l$++)}{
        \For{(z = 0; z $<$ {\tt apr.z\_dim[}$l${\tt ]}; z++)}{
            \For{(x = 0; x $<$ {\tt apr.x\_dim[}$l${\tt ]}; x++)}{
                r\_begin, r\_end $\gets$ {\tt apr}.{\it get\_row}($l, z, x$)\;
                \For{(i = r\_begin; i $<$ r\_end; i++)}{
                    $y \gets$ {\tt apr.y\_idx[}$i${\tt ]}\;
                    $p \gets$ {\tt intensity[}$i${\tt ]}\;
                    \tcc{do something}
                }
            }
        }
    }
    \BlankLine
    \label{alg:linear_iteration}
    \caption{Iteration over particles and coordinates}
\end{algorithm}

Most algorithms require not only iteration over individual particles, but over pairs or neighborhoods of particles. In these cases, the optimal strategy depends on the size and structure of the neighborhood, as well as the type of processing hardware, e.g. CPU or GPU. In the following section we describe the main algorithmic contribution of this work, which includes two iteration strategies for accessing pairs and groups of particles.

\subsection{Neighborhood data access}

Some of the most fundamental operations in image processing, such as spatial filtering and resampling operations, compute values for each pixel in the output image by accumulating values over fixed-size neighborhoods of the input image. These operations can be extended to the APR by instead considering neighborhoods of particles. However, due to the adaptive resolution, the structure of these neighborhoods varies across the domain. This complicates both the definition and implementation of such operations. 

\begin{figure}[h!]
    \centering
    \includegraphics[width=0.45\textwidth]{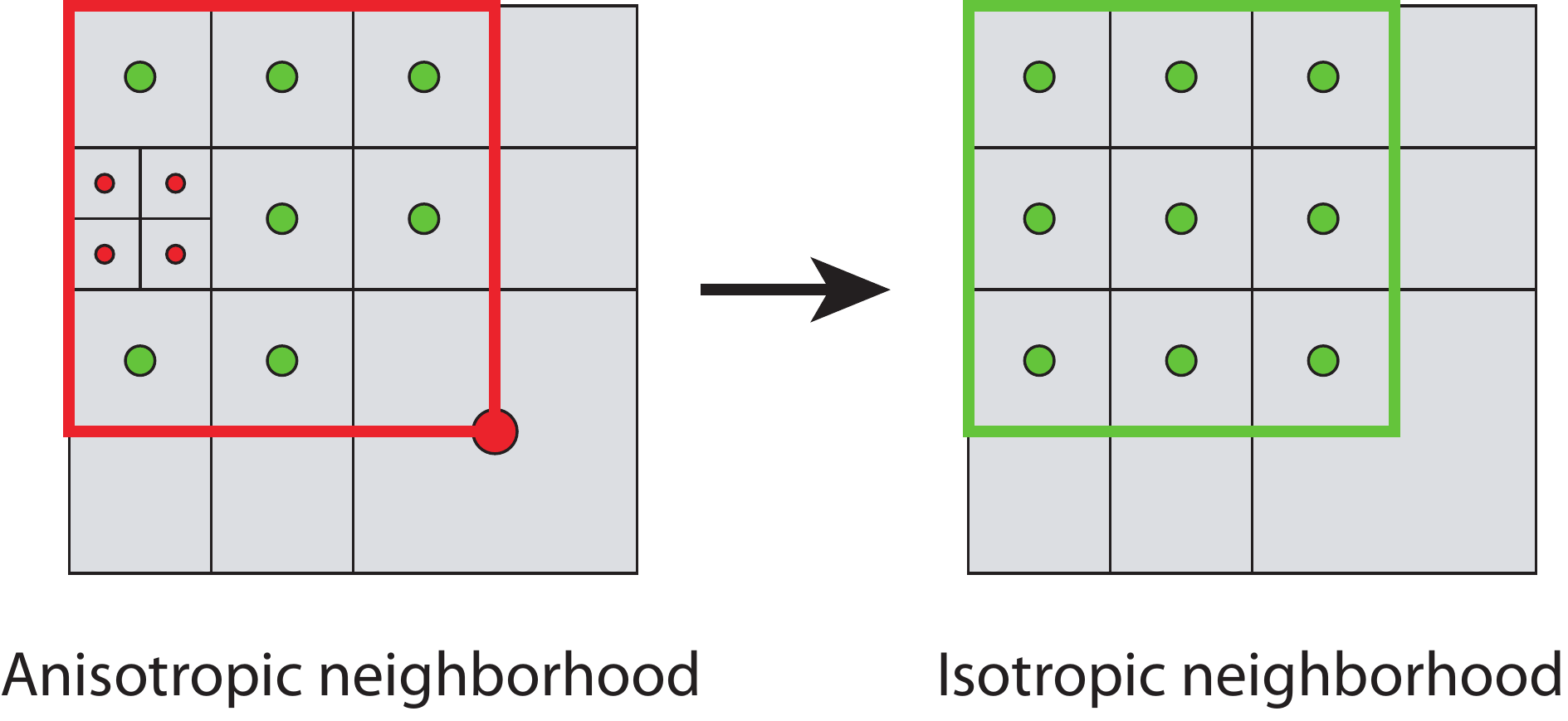}
    \caption{Particle values can be interpolated between resolution levels to create locally isotropic neighborhoods with guaranteed error bounds.}
    \label{fig:isopatch}
\end{figure}

Rather than adapting the operations to anisotropic or non-uniform structures, the reconstruction condition in Eq.~\ref{eq:r_con} enables the interpolation of information across resolution levels. This can be used to interpolate neighborhoods of particles to small local patches of uniform resolution, as shown in Fig~\ref{fig:isopatch}. Indeed, thanks to the reconstruction condition, coarse particles can be interpolated to the finest resolution level with a guaranteed point-wise bound on the interpolation error. This also implies a bounded error at all intermediate resolutions, compared to a similarly downsampled signal, i.e., the corresponding node in the image pyramid (or the full tree in Fig.~\ref{fig:compare}A). Thus, since the APR particles and interior nodes coincide exactly with the corresponding tree nodes and the missing ``pruned'' nodes can be approximated, any image region can be reconstructed at any resolution with guaranteed error bounds.

While conceptually simple, this level interpolation constitutes an algorithmic challenge as it requires iterations over particles in multiple rows, across multiple resolution levels. For example, the $3\times3$ patch reconstruction in Fig.~\ref{fig:isopatch} needs intensity values from 2 sparse rows at the coarser resolution level, 3 sparse rows at the target (medium) resolution, and 6 sparse rows at the finer resolution. Thus, reconstructing such a patch requires iterations, or search, over 11 sparse rows across three levels. In 3D, a corresponding $3\times 3\times 3$ patch uses values from 49 sparse rows. The number of rows required for local patch interpolation grows exponentially with dimension as additional finer resolution levels are included. In other words, reconstructing isotropic patches on the fly, directly from the particles, is not feasible in practice. We address this by introducing an auxiliary data structure, which we call the \emph{APR tree}.

\subsection{The APR tree}

We define the APR tree as the set of all interior nodes in the tree structure, i.e., all yellow nodes in Fig.~\ref{fig:compare}B. Equivalently, this can be thought of as the set of elements in the image pyramid that can be obtained by downsampling the APR particles. We store and access these nodes using a separate instance of the {\tt LinearAccess} APR data structure from Algorithm~\ref{alg:linear_access}. The values of all APR tree particles (interior nodes) can be computed from the APR by recursive reduction over descendant particles. This requires iterating over all APR particles (leaf nodes), while simultaneously accessing the ascendant nodes. Algorithm~\ref{alg:parent_iteration} shows how this can be done, using what we call \emph{synchronized iteration}. For each particle in the leading sparse row, the parent iterator is incremented until the $y$-indices align\footnote{Note that Algorithm~\ref{alg:parent_iteration} depends on the assumption that the parent particle always exists, which for APRs is true by construction. The iteration strategy can be extended to pairs or groups of sparse rows with other spatial relations, but this likely requires additional conditions to be checked.}.

\begin{algorithm}[h!]
    \DontPrintSemicolon
    \KwData{LinearAccess {\tt apr}, LinearAccess {\tt tree}}
    \KwIn{$l$, $z$, $x$}
    \tcc{APR particle row}
    r\_begin, r\_end = {\tt apr}.{\it get\_row}($l, z, x$)\;
    \tcc{Parent row in the APR tree}
    p\_begin, p\_end = {\tt tree}.{\it get\_row}($l-1, z/2, x/2$)\;
    
    $j$ = p\_begin \tcp{parent index}
    \tcc{Iterate over APR particles}
    \For{(i = r\_begin; i $<$ r\_end; i++)}{
        \tcc{Increment j until y-indices align}
        \While{({\tt tree.y\_idx[}$j${\tt ]} $<$ {\tt apr.y\_idx[}$i${\tt ]}$/2$)}{
            $j$++\;
        }
        \tcp{tree node j is the parent of apr node i}
    }
    \label{alg:parent_iteration}
    \caption{Synchronized iteration over a sparse row of particles and their parent nodes in the APR tree}
\end{algorithm}

For separable reduction operators, such as maximum or average reductions, synchronized iteration can be used to compute the values of all particles in an APR tree in the following two steps:
\begin{enumerate}
    \item iterate over all APR particles, reducing their values onto the parent nodes;
    \item iterate over all APR tree particles, level by level from the finest resolution, to fill the tree.
\end{enumerate}
This algorithm can be parallelized by distributing the sparse rows in each step across different threads or processes. However, care must be taken to avoid competing writes to the same address. We do this by distributing blocks of $2\times2$ (in 3D) rows, such that each block corresponds to a single parent row.

\subsection{Isotropic patch reconstruction}\label{sec:patch}

By precomputing the values of all APR tree particles, locally isotropic patches can be reconstructed at any level without iterating over the sparse rows at finer resolutions. This reduces the complexity of this operation. In the following, we assume that all particle values are reduced by average downsampling, and that piecewise constant APR reconstruction is used\footnote{For general linear reduction and reconstruction methods, the interpolation from level $l_0$ to $l_1$, where $l_0 < l_1 < l_{\max}$, can be modified such that it corresponds to interpolation from $l_0$ to $l_{\max}$ followed by downsampling to $l_1$.}. 

We denote the APR particles by $\mathcal{P} = \{(\mathbf{x}_p, f(\mathbf{x}_p))\}_{p=1}^{N_p}$ and the APR tree by $\mathcal{T} = \{(\mathbf{x}_t, f(\mathbf{x}_t))\}_{t=1}^{N_t}$. Suppose we wish to reconstruct the value $\hat{f}(\mathbf{x}_l)$ of a grid cell $\mathbf{x}_l$ at resolution level $l \leq l_{\max}$. There are two possible cases:
\begin{enumerate}
    \item There exists an APR particle $(\mathbf{x}_p, f(\mathbf{x}_p))$ at resolution level $l_p \leq l$ whose particle cell coincides with or contains $\mathbf{x}_l$.
    \item The grid cell $\mathbf{x}_l$
    contains multiple APR particles at finer resolutions $l_p > l$. In this case, there exists a tree particle $(\mathbf{x}_t, f(\mathbf{x}_t))$ at level $l_t = l$ whose cell coincides with $\mathbf{x}_l$.
\end{enumerate}
Since we use piecewise constant reconstruction, we obtain $\hat{f}(\mathbf{x}_l) = f(\mathbf{x}_p)$ in the first case and $\hat{f}(\mathbf{x}_l) = f(\mathbf{x}_t)$ in the second case. Thus, reconstructing an isotropic patch at level $l$ requires iteration over APR particles within the patch at levels $l' \leq l$ and, if $l < l_{\max}$, tree particles at level $l$. The {\tt LinearAccess} data structure from Algorithm \ref{alg:linear_access} allows patches of arbitrary shape and extent in the densely encoded dimensions. However, for performance reasons, it is best to let the patch span the entire sparse $y$-dimension in order to avoid linear searches due to the sequential access.

\section{APR processing}

The ability to locally reconstruct isotropic neighborhoods allows for a wide range of image processing algorithms to be implemented natively for the APR. Arguably the most important low-level vision task is discrete convolution. Not only does this enable spatial filters, but it is also an essential component of high-level vision algorithms including convolutional neural networks. We therefore start by describing APR-native discrete convolution.

\subsection{APR filtering by discrete convolution}

For a pixel image $\mathbf{u} \in \mathbb{R}^\mathbf{s}$ of size $\mathbf{s}\in\mathbb{N}^d$ in dimension $d$, we refer to spatial filtering as the process of applying a dense discrete convolution operation with a stencil $\mathbf{w}\in \mathbb{R}^\mathbf{f}$, $\mathbf{f}\in\mathbb{N}^d$, of the same dimensionality but not necessarily the same size\footnote{Typically, the size of the stencil is on the order of a few pixels in each dimension.} as $\mathbf{u}$.
The output $\mathbf{o} = \mathbf{u} * \mathbf{w}$ produced by this operation is again an image, where each pixel is a linear combination of a neighborhood of pixels in $\mathbf{u}$, weighted by the values in $\mathbf{w}$. Now consider an APR $\mathcal{P}$ representing $\mathbf{u}$. It is certainly possible to approximate $\mathbf{o}$ by reconstructing the pixel image $\hat{\mathbf{u}}$ from $\mathcal{P}$ according to the reconstruction condition and computing $\hat{\mathbf{o}} = \hat{\mathbf{u}} * \mathbf{w}$. The output image $\hat{\mathbf{o}}$ can then again be converted to a new APR $\hat{\mathcal{P}}$ using the pulling scheme. When used in this way, the APR acts as a lossy compression technique, but provides no computational speedup over directly processing $\mathbf{u}$. 

We therefore aim to define a native discrete convolution operation for the APR, which is consistent with convolving a full reconstructed image, but only requires computational operations over the particles. Thus, rather than interpolating coarse particles to the finest resolution, we extend the convolution to operate directly on the coarser resolution levels where appropriate. This can be done by convolving each resolution level separately, similar to how convolutions can be applied to each level of an image pyramid.

\begin{figure}[h!]
    \centering
    \includegraphics[width=0.7\linewidth]{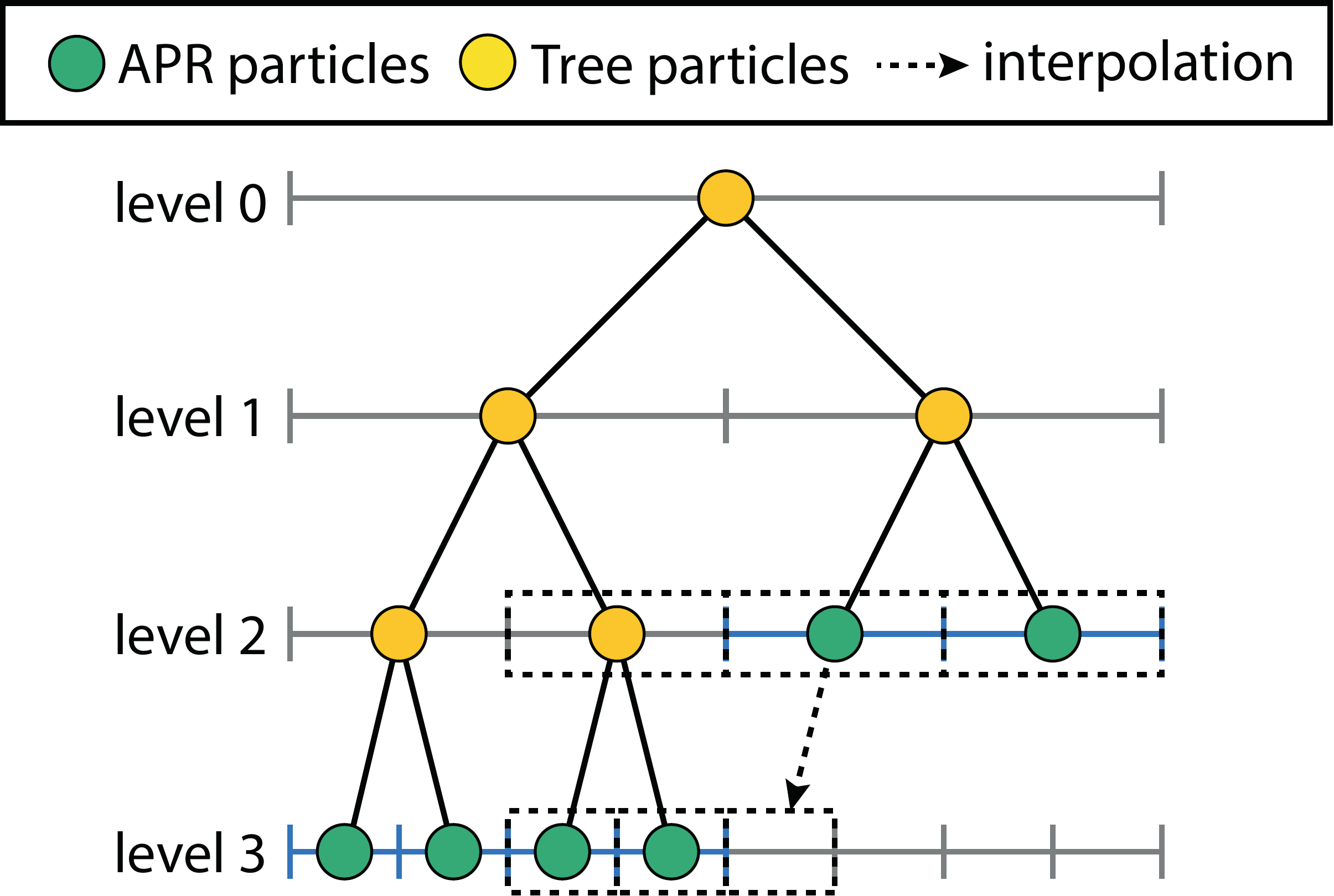}
    \caption{Illustration of APR convolution in one dimension. Filter stencils (dashed boxes) are applied centered on each APR particle (green dots), with neighboring particle values interpolated to the appropriate resolution, to compute the output value at the corresponding center particle. Interpolation from coarse to fine resolution is done on the fly, while values interpolated from fine to coarse resolution are precomputed and accessed via the auxiliary APR tree structure (yellow dots).}
    \label{fig:treeconv}
\end{figure}

An APR corresponds to a sparse disjoint sampling of an image pyramid constructed from the original pixel image. Moreover, isotropic patch reconstruction (cf.~Section \ref{sec:patch}) allows us to approximate any region in this pyramid with a theoretically bounded reconstruction error according to Eq.~\ref{eq:r_con}. This affords some freedom in designing the convolution operation. Here, we define a convolution as an operator that acts only on the particle values $f(\mathbf{x}_p)$, keeping their positions $\mathbf{x}_p$ fixed. More precisely, the filter is applied centered on each particle to compute the output value at that location. Neighboring particles are interpolated to an isotropic patch at the resolution of the center particle. The size of the patch must be at least the size of the stencil $\mathbf{w}$. This is illustrated in Fig.~\ref{fig:treeconv}. In practice, it is typically advantageous to reconstruct larger patches such that the stencil can be applied to multiple locations in each patch.

The APR convolution, as defined here, can be thought of as a feature extractor which computes a feature at each particle location by accumulating neighborhood information. However, in general one cannot guarantee that the resulting particle values, at the sampling locations computed from the input signal $\mathbf{u}$, is a close approximation to the filtered signal $\mathbf{o}$. This is easily seen by considering, for example, a filter which shifts the signal $k>0$ pixels in a given direction. In order to properly represent the output of this operation, the resolution function would have to be shifted similarly. Here we assume that the sampling locations, and thus the resolution function, remain fixed. Thus, the result of the APR convolution can only be viewed as an approximation of $\mathbf{o}$ if the effect of the filter is such that it does not increase the required resolution at any particle location. This is a rather strong condition, which does not hold for arbitrary filters. In particular, it excludes filters that result in a significant net translation of the signal. But it holds, at least approximately, for important image filters such as gradient and sharpening, as well as near-symmetric low-pass (smoothing) filters.

At the finest resolution level, the APR convolution operation is equivalent to convolving the reconstructed image $\hat{\mathbf{u}}$. With decreasing resolution the spatial extent of the filter grows, and the convolution is applied to (approximations of) downsampled versions of the image. Thus, applying the same stencil $\mathbf{w}$ at coarser resolution levels is, in general, not consistent with the convolution at the finest resolution. We address this by modifying the stencil depending on the resolution level.

\subsection{Pixel-consistent APR convolutions by stencil restriction}

In order to render APR convolution consistent with a convolution of the fully reconstructed image $\hat{\mathbf{u}}$, we use \emph{operator restriction} as commonly encountered in the algebraic multigrid literature \cite{haber2018learning,trottenberg2000multigrid}. Denote by $R_l$ the restriction (downsampling) operator used to determine particle values at a coarse resolution level $l < l_{\max}$, and let $P_l$ be the prolongation (upsampling) operator to reconstruct pixel values from particles at level $l$. Let $\hat{\mathbf{u}}_l$ be the image reconstructed at level $l$ and $\hat{\mathbf{u}}$ the finest reconstruction at pixel resolution. We then have:
\begin{equation}
    \hat{\mathbf{u}}_l = R_l \hat{\mathbf{u}}\, , \quad \hat{\mathbf{u}} \approx P_l \hat{\mathbf{u}}_l \, .
\end{equation}
That is, $\hat{\mathbf{u}}_l$ can be computed from $\hat{\mathbf{u}}$ by downsampling via $R_l$, and $\hat{\mathbf{u}}$ can be approximated by upsampling via $P_l$. Now consider the convolution of $\hat{\mathbf{u}}$ with a stencil $\mathbf{w}$. This is a linear operation, which can be written as a matrix multiplication $\mathbf{w} * \hat{\mathbf{u}} := K_w \hat{\mathbf{u}}$. Using the interpolation operators $R_l$ and $P_l$, we can now define a convolution operator $K_{w_l}$ for the coarse image $\hat{\mathbf{u}}_l$ on resolution level $l$ as
\begin{equation}
    \label{eq:restrict_stenc}
    K_{w_l} \hat{\mathbf{u}}_l := R_l K_w P_l \hat{\mathbf{u}}_l.
\end{equation}
In this way, $K_{w_l}$ is equivalent to interpolating $\hat{\mathbf{u}}_l$ to the fine resolution, applying the convolution with $\mathbf{w}$ there, and downsampling the result back to the coarse resolution. Similar to $K_w$, the coarse scale Toeplitz matrix $K_{w_l}$ corresponds to a stencil $\mathbf{w}_l$, whose structure depends on the fine resolution stencil $\mathbf{w}$, as well as the restriction and prolongation operators. For stencils $\mathbf{w}$ that are spatially invariant, the stencil $\mathbf{w}_l$ can be computed without generating the matrix $K_{w_l}$, by evaluating only the $|\mathbf{w}_l |$ unique non-zero elements.

The interpolation operators used in the standard APR formulation, i.e., block-wise average downsampling and piecewise constant upsampling, amount to particularly simple forms of $R_l$ and $P_l$. This combination additionally has the property that $R_l P_l = I$, where $I$ is the identity matrix. In this case, the restricted stencil $\mathbf{w}_l$ is computed directly by averaging the contributions of all elements of the fine stencil $\mathbf{w}$ applied at all possible positions within the coarse grid cell. 

This provides for internally consistent APR convolution by restricting the filter stencil to each resolution level, of which there are at most $\lceil\log_2(|\Omega|)\rceil$, where $|\Omega|$ is the maximum image side length in pixels.
Figure~\ref{fig:fmd} shows an example of this approach for a Gaussian smoothing filter (middle row) applied directly on the APR levels after restriction (right panel), compared with applying the same filter on the raw pixels (middle panel) and on a noise-free ground-truth (GT) version of the pixel image (left panel). The respective input images are shown in the top row of the figure. The results of the smoothing filter computed on raw pixels and on the APR are visually indistinguishable and have similar PSNR and SSIM with respect to the blurred GT image. This confirms that the restricted APR filters provide pixel-consistent results.

\subsection{Resolution-adaptive filters}

\begin{figure*}[htb]
    \centering
    \includegraphics[height=0.5\linewidth]{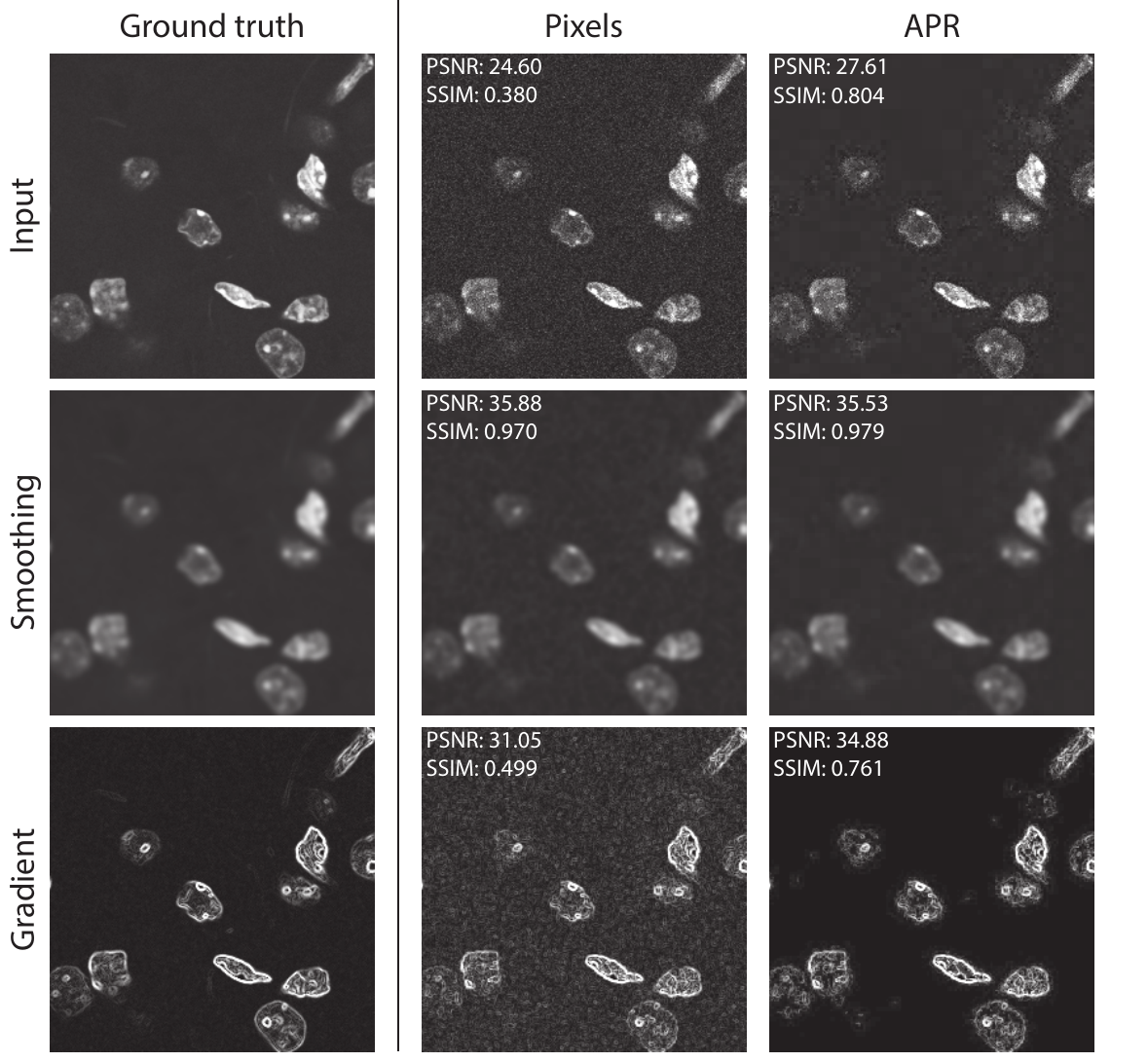}
    \caption{Illustration of APR filtering for smoothing and gradient estimation. The images are taken from the FMD dataset\cite{zhang2018poisson}. Top row: ground truth (GT) image obtained by averaging 50 acquisitions, raw image (a single acquisition), and the APR of the raw image. Middle row: Gaussian smoothing with standard deviation $\sigma=2$ pixels applied to each image from the top row, using restricted stencils for coarser APR particles. Bottom row: gradient magnitudes computed from the respective image of the top row. The ground truth gradients were computed using central finite differences, whereas the gradients on noisy raw image were computed using Sobel filters. For the APR, the filters were rescaled according to the particle distance at each resolution level (see Eq.~\ref{eq:adapt_grad}). All PSNR and SSIM values are given with respect to the GT image in the same row.\label{fig:fmd}}
\end{figure*}

The APR convolution operation is inherently spatially adaptive, as guided by the structure of the APR. In the previous section, we adjusted the convolution stencils on each APR level such that the APR convolution is a consistent approximation of the full-pixel convolution. However, this may not always be the goal. Indeed, the spatial adaptivity of the APR can be exploited to provide scale-adaptive image filters. This can improve filter performance, e.g., in image gradient estimation. Since the smoothing length of the gradient estimator depends on the spacing between neighboring discretization points, it naturally varies with the APR resolution level. Thus, APR convolution with a level-dependent stencil $\mathbf{w}_l$ of the form
\begin{equation}
    \mathbf{w}_l = 2^{-(l_{\max}-l)} \mathbf{w},
    \label{eq:adapt_grad}
\end{equation}
where $\mathbf{w}$ is the pixel-level stencil, naturally adapts gradient estimation to varying length scales in the image contents. The benefits are illustrated in Fig.~\ref{fig:fmd} (bottom row) for gradients estimated using a Sobel filter that combines central finite differences with a smoothing filter orthogonal to the gradient direction. When used on the APR with level-dependent rescaling according to Eq.~\ref{eq:adapt_grad}, the gradients along the edges of the objects are captured, while being less sensitive to imaging noise in low-gradient regions. This is reflected in the higher PSNR and SSIM values of the result computed on the APR compared to the result computed on pixels.

\section{Parallel implementation}

The APR convolution operation described so far essentially consists of three algorithmic components:
\begin{enumerate}
    \item computing the values of (interior) APR tree particles,
    \item reconstructing isotropic patches, and
    \item accumulating patch values to form the convolution output.
\end{enumerate}
Filling the APR tree (step 1) is algorithmically independent of the convolution operation, while steps 2 and 3 are performed alternately for each location in the image domain.
In order to take advantage of modern computing hardware, we have implemented these algorithmic components to use thread parallelism on multi-core CPUs and on GPUs. We describe the design principles used to achieve this and discuss their advantages and limitations.

\subsection{CPU parallelization using OpenMP}

Modern CPUs typically comprise 6 to 48 cores, each optimized to execute a pipelined sequence of instructions called a \emph{thread}. Parallelization on multi-core CPUs thus amounts to dividing a program into a small number of threads that can be distributed across the cores and executed concurrently. We provide a thread-parallel CPU implementation of the present APR convolution algorithm based on the Open Multi-Processing (OpenMP) \cite{dagum1998openmp} API, which provides a set of high-level directives and work-sharing constructs for SPMD (Single Program Multiple Data) parallelism, where the same set of instructions is executed concurrently at multiple data points. 

In order to distribute the data points, in our case APR particles, across threads, we follow a domain-decomposition approach as is classic for pixel images. Execution efficiency demands that the threads iterate over the sparse $y$-dimension, which is accessed sequentially according to Algorithm~\ref{alg:linear_iteration}. Therefore, we partition the APR by assigning to each CPU core a subset of the sparse rows, rather than a subset of the particles. However, the number of particles in each sparse row can vary greatly. This may lead to large differences in the amount of work done by each core, causing slowdown due to load imbalance. We address this by runtime dynamic load balancing, dynamically re-distributing sparse rows between threads at runtime. 

Consider the convolution with a stencil $\mathbf{w}$ of size $k^3$ at some resolution level $l$ in the APR of an image of size $N_z \times N_x \times N_y$ pixels. The data are distributed such that each thread computes the output for the APR particles in a given $z$-slice. Each thread allocates a buffer of size $k \times k \times (N_y + k-1)$ for the isotropic patches, where the $k-1$ additional points in the $y$-dimension allow for padding to handle boundary conditions at the edges of the image. For each $z$-slice, the buffer is iterated over along the $x$-dimension with the center of the buffer at the given $z$-index. At each location, the thread iterates over the necessary\footnote{Isotropic patch reconstruction does not have to include the particles from all coarser levels, but enough to ensure that neighborhoods of size $k^3$ around each output particle are filled. Since the resolution levels of neighboring particles cannot differ by more than one~\cite{cheeseman2018adaptive}, the number of coarser levels that must be considered scales logarithmically with the stencil radius $(k-1)/2$.} sparse rows to fill the $k$ new rows at the beginning of the buffer (cf.~Section \ref{sec:patch}). The output values are then aggregated for each APR particle in the sparse row corresponding to the center location of the buffer.

\subsection{GPU parallelization using CUDA}

In addition to the multi-threaded CPU implementation, we also provide an optimized implementation of the present APR convolution algorithm for GPUs. Our implementation uses the Compute Unified Device Architecture (CUDA) API for GPU programming and is hence limited to Nvidia GPUs. Recent Nvidia GPUs boast thousands of cores organized into arrays of \emph{Streaming Multiprocessors} (SMs), each designed to concurrently execute hundreds of threads. In order to parallelize a program using CUDA, it is divided into \emph{thread blocks} (TBs). Each TB executes a program, called a \emph{CUDA kernel}, over a given block of data. The threads within a TB are executed on the same SM in groups of 32 threads, called \emph{warps}. Threads within the same TB communicate via shared memory and barrier synchronization, with recent versions of CUDA additionally supporting warp-level communication and synchronization.

CUDA programs that make good use of the hardware resources of a GPU need to adhere to multiple design principles: First, all threads within a warp should simultaneously perform the same operations. This maximizes efficiency because the threads in a warp execute instructions in lockstep following to the SIMT (Single Instruction Multiple Threads) model of parallelism. Second, diverging control flow paths within a warp are to be avoided. Otherwise, the alternative control flow paths are executed sequentially, with masking employed to enable and disable the appropriate threads. Third, access to the global memory has to be minimized and done in stride, as those transactions carry a significant overhead. This is mainly achieved by \emph{latency hiding}, where execution switches to another warp while one warp waits for the memory. Latency hiding requires that enough scheduled threads are available, providing a fourth design requirement. Fifth, memory latency can be further reduced by: (1) loading frequently accessed data into shared memory or registers and (2) ensuring that the necessary global memory accesses are coalesced, such that each transaction serves as many threads as possible at once. 

These programming requirements are not easily reconciled with a dynamically adaptive tree data structure like the APR. Considering the APR data structures proposed in Algorithm~\ref{alg:linear_access},  particle values and $y$-indices are contiguous in memory along the sparse $y$-dimension. In order to promote coalesced memory access to these vectors, adjacent threads in a warp should therefore access adjacent particles within the same sparse row. We achieve this by distributing the sparse rows across warps. The size of each thread block is limited to 1024 threads on most GPU devices. Since a warp consists of 32 threads, one warp must handle multiple sparse rows if an operation requires accessing more than 32 different positions in $z$ and $x$. This may lead to reduced concurrency due to increased register usage. By instead assigning a half-warp (16 threads) to each sparse row, this limitation can be relaxed to allow up to 64 different positions. Therefore, our design imposes an upper limit to the size of the stencil neighborhood that can be accessed during convolution. Assuming a cube-shaped stencil in 3D, with odd side length, the size is limited to $7^3$ pixels.

\begin{figure}[h]
    \centering
    \includegraphics[width=\linewidth]{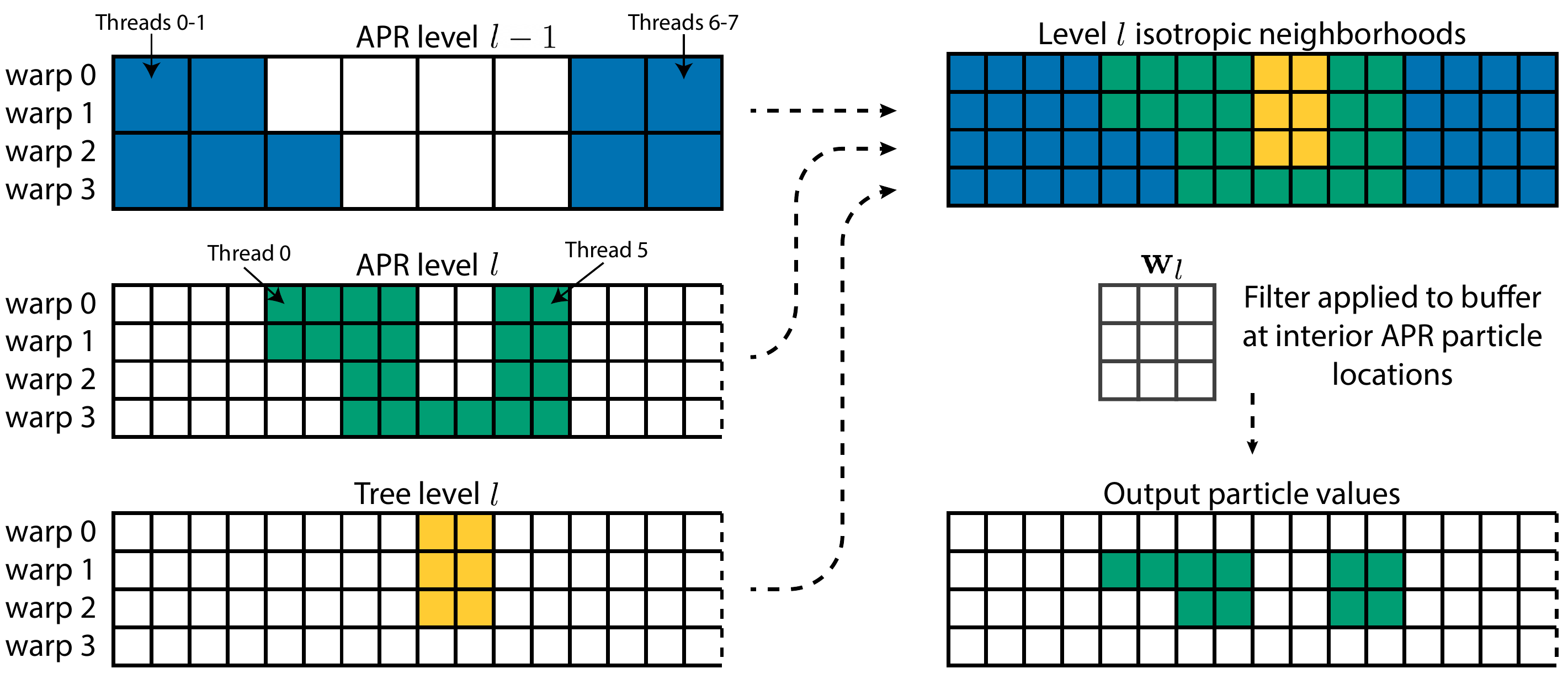}
    \caption{Illustration of our CUDA kernel operations for APR convolution at an arbitrary APR level $l$ in 2D. Each warp 
    corresponds to a fixed position $(z, x)$ at level $l$. The threads load particles from up to three sparse rows at the corresponding position: the APR at levels $l-1$ (top-left panel) and $l$ (middle-left), and the APR tree at level $l$ (bottom-left). Particle values are written to a uniform shared-memory buffer at level $l$ to reconstruct local isotropic neighborhoods around target particle locations (top-right, loaded particle origins identified by color). Finally, the convolution stencil $\mathbf{w}_l$ 
    (middle-right) is applied to interior positions in the buffer, computing the output at the level $l$ APR particle locations (bottom-right).}
    \label{fig:warps}
\end{figure}

Using this design, we provide optimized implementations of 3D APR convolutions in CUDA for filter stencils of size $3^3$ and $5^3$ pixels. For the sake of example, we describe the $3^3$ algorithm, assuming that the tree particle values $f(\mathbf{x}_t)$ of $\mathcal{T}$ have previously been computed. For this stencil size, we choose a thread block size of $32\times4\times4$, resulting in 16 warps at different positions in a $4\times4$ patch in the $z-x$ plane. Isotropic patches are reconstructed in a shared memory buffer of size $4\times4\times32$. Reconstructing neighborhoods of size $3^3$ around APR particles at level $l$ requires values from the APR at levels $l-1$ (if $l>l_\mathrm{min}$) and $l$, as well as the APR tree at level $l$ (if $l<l_\mathrm{max}$). Thus, each thread accesses particles from up to three sparse rows, one from each of these structures, corresponding to the fixed $(z, x)$ position of the warp. The shared memory buffer is iterated across the $y$-dimension in a synchronized manner. For each position of the buffer, the threads:
\begin{enumerate}
	\item Update their particles by comparing the $y$-indices to the current location of the buffer. If a particle is behind the buffer (i.e. has a $y$-index smaller than the beginning of the buffer), the thread loads the next particle in the corresponding row. This is repeated until no particles are behind the buffer any more.
	\item Reconstruct local isotropic patches by writing particle values within the range of the patch to the shared memory buffer.
	\item Aggregate the convolution output for the level $l$ APR particles in the $2\times2$ center rows of the buffer.
\end{enumerate}
Block-level barrier synchronization is required before steps 2 and 3 to ensure correctness of the output. Figure~\ref{fig:warps} illustrates this algorithm in 2D. The implementation for stencils of size $5^3$ is analogous, with the main difference being the size of the thread block, which we then choose to be $16\times8\times8$.

We additionally implement high-level optimization strategies to reduce unnecessary work for sparse images. In this case, the sparse rows at fine resolution levels tend to be empty or contain only few particles. Therefore, thread blocks may needlessly reconstruct patches where there are no target particles. We avoid this by using a pre-processing CUDA kernel to detect the positions in the APR where the sparse rows contain at least one particle. The convolution stencils are then launched only at those locations. In addition, the thread blocks use a reduction operation to determine the minimum and maximum $y$-indices of the target particles. Positions of the uniform buffer outside of this range are skipped.

\section{Results}

We present benchmark results of computational performance and memory usage of our APR convolution implementations using both OpenMP on CPUs and CUDA on GPUs. We also present an example of image deconvolution using the Richardson-Lucy algorithm re-implemented for APRs and compare the results with those obtained by the classic algorithm on pixels. 

Since the APR adapts to the contents of an image, the performance of APR processing algorithms depends on image contents through the number of particles required by the APR to reach the reconstruction error bound. We quantify this using the \emph{computational ratio} (CR), defined as
\begin{equation}
    \text{CR} = \frac{\text{Number of pixels in the original image}}{\text{Number of particles in the APR}}.
\end{equation}
To put this metric into perspective, typical CR values for real-world fluorescence microscopy datasets range from 3.6 to 372.2 with an average of 51.1 and a median of 22.7~\cite{cheeseman2018adaptive}. The CR value one can obtain for any given image depends mainly on the sparsity of the image content. The APR will represent regions where the signal gradient is significant (relative to the local error scale $\sigma$) at high resolution. Thus, if the gradients are distributed densely across most of the image, high CR values cannot be obtained without significant loss of information.

In addition to the CR, also other factors may directly or indirectly affect memory usage and computational performance. For example, the size of the APR data structure explicitly depends on the original image size in the $x$ and $z$ dimensions. Moreover, performance may vary between different APRs, even if both the image dimensions and the CR are equal, since the particle layout can affect, e.g., cache hit rates and conditional branches.

\subsection{APR convolution benchmarks} \label{sec:benchmarks}

Here we present results on the memory usage and computational performance of our APR convolution implementations. The benchmarks are performed on APRs computed from a set of $10$ synthetic images, containing spherical objects at varying densities, yielding a range of CR values from 1.04 to 1019.8. The files, as well as the code used to produce all data presented here, are available in LibAPR.

\subsubsection{Memory usage}

In order to convolve an image of $N$ pixels, the simplest approach is to allocate enough memory to hold the entire image plus an output buffer of the same size. If the input and output data types require $s_i$ and $s_o$ bytes of storage, respectively, this requires $N (s_i + s_o)$ bytes of memory. Similarly, convolving an APR requires two buffers for the input and output particle intensities, but in addition also the access data structures of the APR and the APR tree, as well as the interior particle intensity values.

We benchmark this using synthetic images of size $1024^3$ pixels and of different CR. The benchmark images are described and examples shown in Appendix~\ref{supp:images}. Figure~\ref{fig:cudaconv_mem} shows a breakdown of the memory requirements for APR convolution when all pixel and particle intensity values are stored as 32-bit data types. The straightforward convolution on pixels requires 8.59\,GB of memory\footnote{Throughout this manuscript, we use decimal SI prefixes for Byte multiples, so for example 1\,GB = $10^9$ Byte.}, exceeding the available VRAM of most modern GPUs. Tiling strategies must then be implemented to enable processing, but this is only efficient if image tiles can be transferred to and from the GPU device concurrently with computation. This adds complexity to the implementation and most likely results in a performance penalty. Computing the convolution natively on an APR with a CR of 20.8 (which is smaller than the median of real-world microscopy datasets~\cite{cheeseman2018adaptive}) requires 0.58\,GB of memory: 14.8 times less than the equivalent pixel operation. This makes straightforward processing of the image possible on almost any currently available GPU.

\begin{figure}[!htb]
    \centering
    \includegraphics[width=0.95\linewidth]{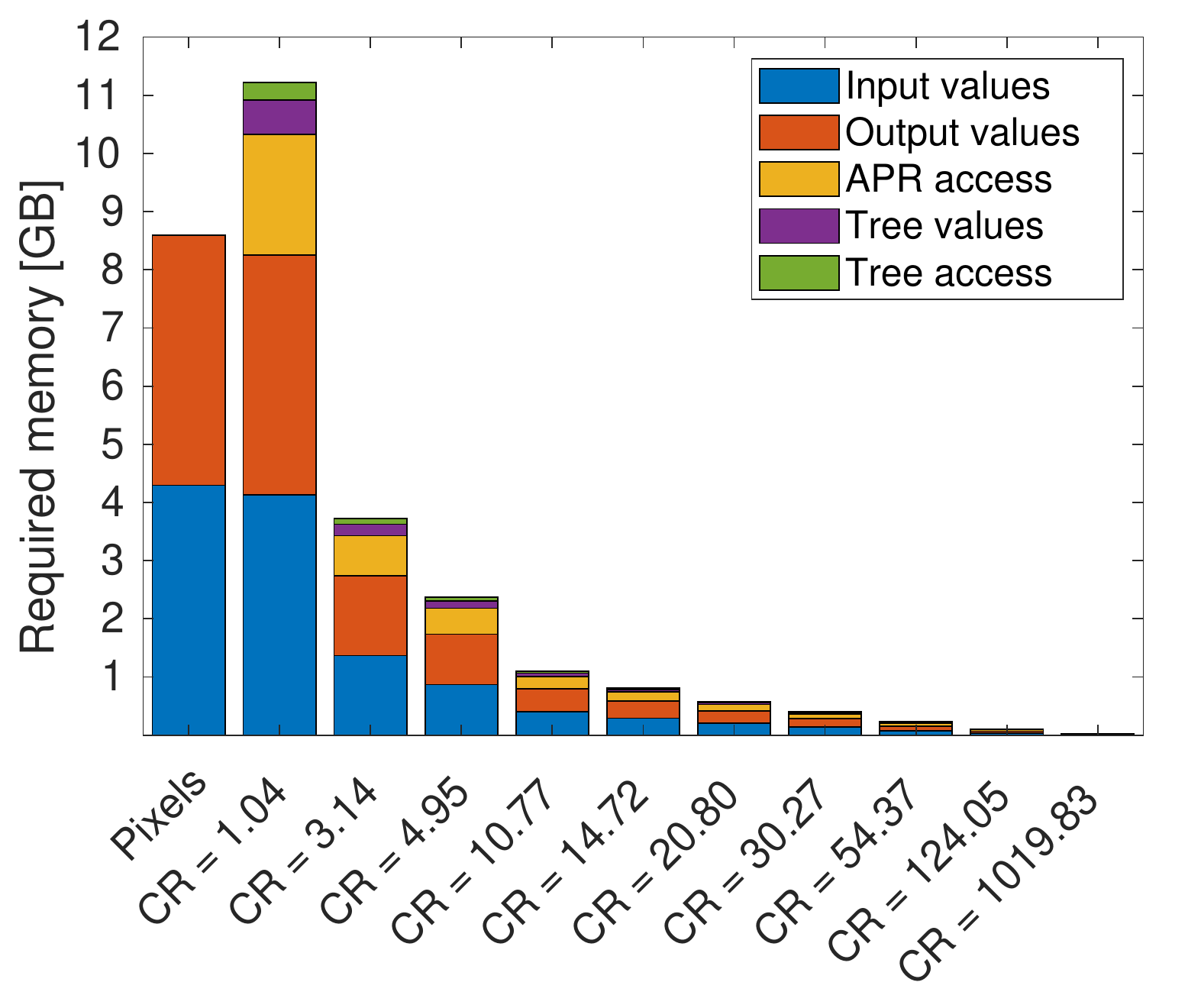}
    \caption{Memory required to perform one convolution operation on a cubic image of $1024^3$ pixels and the corresponding APRs at different computational ratios (CR). Pixel and particle intensities are stored using 32-bit data types. We neglect the memory required to store the weights of the filter stencil. The APR memory usage additionally includes the APR access data structure, as well as the access data structure and values of the APR tree (interior nodes).}
    \label{fig:cudaconv_mem}
\end{figure}

The memory required for APR convolution decreases monotonically with increasing CR, as shown in Fig.~\ref{fig:cudaconv_mem}. For CR$\approx$1, we observe that the memory overhead from the APR data structures is about 30\%. However, this overhead is quickly amortized for higher CRs, since the size of the {\tt xz\_end} vector in the {\tt LinearAccess} data structure is constant (see Algorithm~\ref{alg:linear_access}). APR-native convolution at CR$\approx$1020, a value not uncommon in fluorescence microscopy, requires 25.5\,MB of memory, or 337 times less than the classic pixel-based implementation. 

For GPU implementations, the benefit of reducing the memory requirement is two-fold: First, it enables larger image regions to be kept in memory and processed without tiling strategies. Second, the total amount of data that needs to be transferred to and from the GPU is reduced proportional to the CR, alleviating the performance bottleneck from the host-device transfer bandwidth. 

\subsubsection{Computational performance}

We present benchmark results for APR convolution using filter stencils of $3^3$ and $5^3$ pixels, respectively. All benchmarks are performed on an Alienware m15 laptop equipped with a GeForce RTX 2080 Max-Q GPU and an Intel Core i9-9980HK CPU running Ubuntu 18.04 and CUDA Toolkit 11.0 RC.

The APR convolution implementation in CUDA consists of three steps: 1) finding the locations of non-empty sparse rows\footnote{This is not strictly necessary, but we have found that it yields a speedup of up to a factor of 5 for high CR values, while the additional overhead is negligible for low CR values.}, 2) filling the APR tree, and 3) performing the convolution operation. All of these steps are implemented in CUDA and performed in sequence. In the benchmark results below, all three steps are included in the reported times. However, in practical situations requiring repeated convolutions, step 1 can be reused, and for convolution of the same input with multiple filters, the tree data can also be reused, leading to runtimes smaller than those reported here. 

\begin{figure}[!htb]
    \centering
    \includegraphics[width=0.95\linewidth]{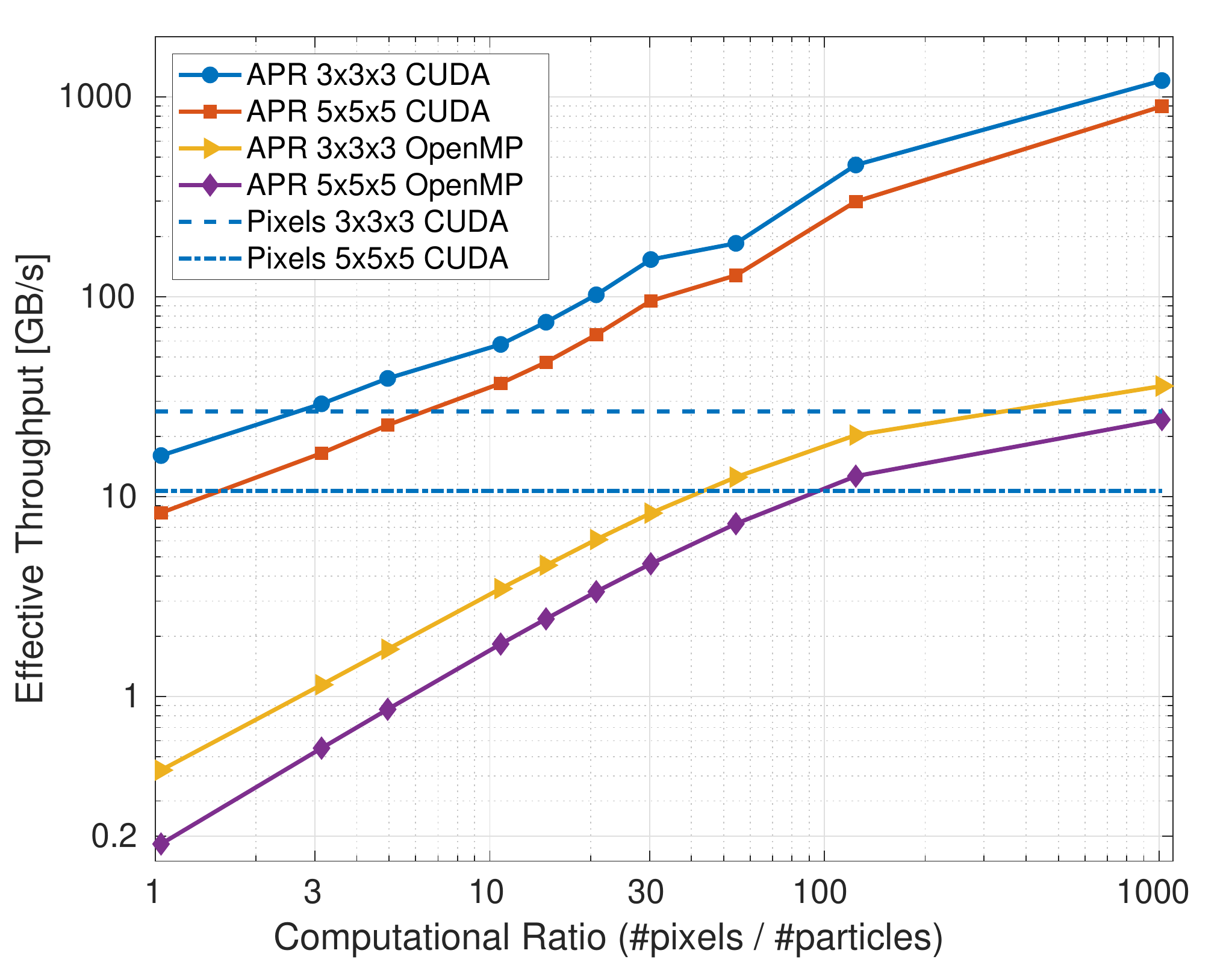}
    \caption{Computational performance of APR convolutions using GPU (CUDA) and CPU (OpenMP) parallelization for $10$ synthetic benchmark datasets of different computational ratios (CR). The effective throughput is calculated as the size of the original image in GB divided by the processing time in seconds. GPU processing times include all computations, but exclude data transfers. The horizontal lines show measurements for pixel convolutions using CUDA ArrayFire. We show results for $3^3$ and $5^3$ stencils.}
    \label{fig:cudaconv_tp}
\end{figure}

Rather than reporting absolute wall-clock times, we put the results on a more intuitive scale by computing the \emph{effective throughput}, which we define as
\begin{equation}
    \mbox{Effective throughput} = \frac{\mbox{size of pixel image in Bytes}}{\mbox{total processing time}}.
    \label{eq:effective_throughput}
\end{equation}
For a pixel algorithm, this is the classic data throughput, i.e., the number of Bytes processed per second. For an APR, the effective throughput states the throughput a pixel algorithm would need to have in order to achieve the same processing time. Figure~\ref{fig:cudaconv_tp} shows the scaling of the effective throughput for the same synthetic images of different CRs as already used in Fig.~\ref{fig:cudaconv_mem}, subsampled to size $512^3$. All data are represented as 32-bit floating point numbers. Hence, the numerator in Eq.~\ref{eq:effective_throughput} is 537 MB. The GPU timings include all computational steps, but exclude the times for host-device data transfers. The runtimes for the corresponding pixel convolutions, performed using the CUDA backend of ArrayFire v3.8.0~\cite{Yalamanchili2015} in C++, are shown as horizontal lines, as they do not depend on the CR. 

The effective throughput of the ArrayFire CUDA implementation is 26.6\,GB/s for the $3^3$ filter and 10.7\,GB/s for the larger $5^5$ filter. Our CUDA APR convolutions break even with the performance of ArrayFire at CR values below $3$. For higher computational intensity of the operation (i.e., for larger stencils), APR convolution breaks even earlier. At a CR of 20.8, the $3^3$ and $5^5$ APR convolution speeds correspond to pixel throughputs of 102.2 and 64.5\,GB/s, respectively. That is 3.8 and 6.0 times faster than the ArrayFire implementation. For the benchmark image with CR=124, which is well within the range typical of real-world microscopy images~\cite{cheeseman2018adaptive}, the effective throughput of APR convolution increases to 455.7 and 299.1\,GB/s, respectively. Compared to ArrayFire, this corresponds to speedup factors of 17 and 28. Considering the theoretical peak performance of 6.447\,TFLOPS (trillion floating point operations per second) of the benchmark GPU, the theoretical performance limit for the pixel convolution is 486\,GB/s for the $3^3$ filter and 103\,GB/s for $5^5$. The effective throughput of the $5^3$ APR convolution exceeds this limit for CR values larger than 30.

The performance of the APR convolution on the multi-core CPU using OpenMP also scales with CR. For large CR values $>100\ldots 500$, APR performance on the CPU even exceeds pixel performance on the GPU. If data transfer times to and from the GPU are taken into account, the CPU breaks even at lower CR values. Comparing the performance of the APR convolution on the GPU and CPU, we observe that the GPU implementation (without data transfer times) is 15 to 45 times faster across benchmark datasets and filter sizes. This is discussed in more detail in Appendix~\ref{supp:benchmarks}. 

In summary, these benchmarks suggest that native APR convolutions are suited to overcome some of the difficulties associated with processing large images on the GPU, providing real-time convolution implementations for edge computing, or accelerating CPU implementations to achieve GPU-like performance.

\subsection{Richardson-Lucy deconvolution} \label{sec:deconv}

We demonstrate an application of native APR convolution for image restoration using the iterative Richardson-Lucy (RL) deconvolution algorithm~\cite{richardson1972bayesian, lucy1974iterative}. Deconvolution is a frequent example of an ill-posed inverse problem in microscopy image processing. Suppose we have acquired a blurred (from light diffraction) and noisy (detector shot noise and electronics noise) image
\begin{equation}
    \mathbf{u} = \eta (\mathbf{i} * \mathbf{w}),
\end{equation}
where $\mathbf{i}$ is the imaged sample (e.g., the spatial distribution of fluorophores in the specimen), $\mathbf{w}$ is the point-spread function (PSF) of the microscope optics, and $\eta$ models the noise distribution. Under the assumption of Poisson-distributed noise, the RL algorithm attempts to recover $\mathbf{i}$ from $\mathbf{u}$ via the iterative updates
\begin{align}
    \mathbf{i}_{k+1} &= \mathbf{i}_k  \left(\frac{\mathbf{u}}{\mathbf{i}_k * \mathbf{w}} * \mathbf{w}^\dagger\right),
\end{align}
which amounts to a fixed-point iteration for maximizing the likelihood of observing $\mathbf{u}$. 
Division and multiplication are element-wise, $*$ denotes discrete convolution, and $\mathbf{w}^\dagger$ is the flipped PSF with the order of elements reversed in each dimension. We adapt this algorithm to the APR by replacing the convolution operations with their APR-native counterparts and restricting the PSF stencil to the different APR levels as given in Eq.~\ref{eq:restrict_stenc}.

\begin{figure*}[htb!]
    \centering
    \begin{tabular}{cccc}
        \textbf{Ground truth} & \textbf{Blurred \& noisy} & \textbf{Deconvolved (pixels)} & \textbf{Deconvolved (APR)} 
        \\
        {
        \centering
        \begin{overpic}[width=0.16\linewidth]{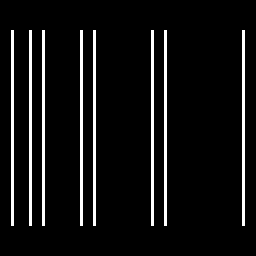}
            \put(2, 98){\color{white}\vector(0,-1){10}} 
            \put(0, 83){\color{white} \scriptsize \textbf{y}}
            \put(2, 98){\color{white}\vector(1,0){10}}  
            \put(13, 95){\color{white} \scriptsize \textbf{x}}
        \end{overpic}
        }
        &
        {
        \centering
        \includegraphics[width=0.16\linewidth]{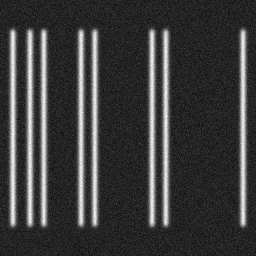}
        }
        &
        {
        \centering
        \includegraphics[width=0.16\linewidth]{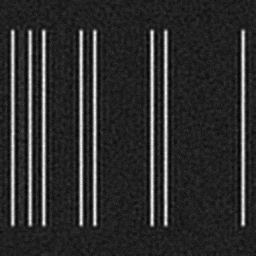}
        }
        &
        {
        \centering
        \includegraphics[width=0.16\linewidth]{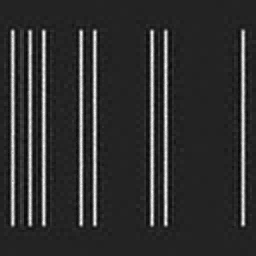}
        }
        \\
        {
        \centering
        \begin{overpic}[width=0.16\linewidth]{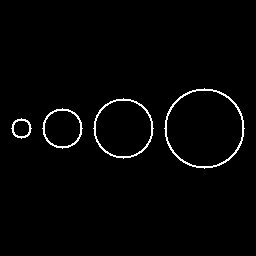}
            \put(2, 98){\color{white}\vector(0,-1){10}} 
            \put(0, 83){\color{white} \scriptsize \textbf{z}}
            \put(2, 98){\color{white}\vector(1,0){10}}  
            \put(13, 95){\color{white} \scriptsize \textbf{x}}
        \end{overpic}
        }
        &
        {
        \centering
        \begin{overpic}[width=0.16\linewidth]{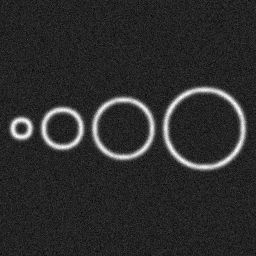}
            \put(1,10){\color{white} \scriptsize \textbf{PSNR: 21.04}}
            \put(1,1){\color{white}  \scriptsize \textbf{SSIM:\, 0.716}}
        \end{overpic}
        }
        &
        {
        \centering
        \begin{overpic}[width=0.16\linewidth]{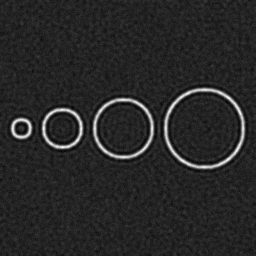}
            \put(1,10){\color{white} \scriptsize \textbf{PSNR: 23.67}}
            \put(1,1){\color{white}  \scriptsize \textbf{SSIM:\, 0.594}}
        \end{overpic}
        }
        &
        {
        \centering
        \begin{overpic}[width=0.16\linewidth]{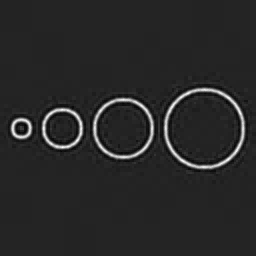}
            \put(1,10){\color{white} \scriptsize \textbf{PSNR: 24.13}}
            \put(1,1){\color{white}  \scriptsize \textbf{SSIM:\, 0.916}}
        \end{overpic}
        }
    \end{tabular}
    \caption{Richardson-Lucy deconvolution of a synthetic image of hollow cylinders. The top row shows the center $xy$-plane ($z=128$) while the bottom row shows the center $xz$-plane ($y=128$). From left to right, we show: 1) the synthetic ground truth image, 2) the image after blurring with a Gaussian filter ($\sigma=2$ pixels, support $13^3$ pixels) and addition of pixel-wise independent additive Gaussian noise, 3) deconvolved images after 100 Richardson-Lucy iterations on the computed on pixels or directly on the APR. The PSNR and SSIM values are computed for the entire volumes ($256^3$ pixels).}
    \label{fig:deconvolution}
\end{figure*}
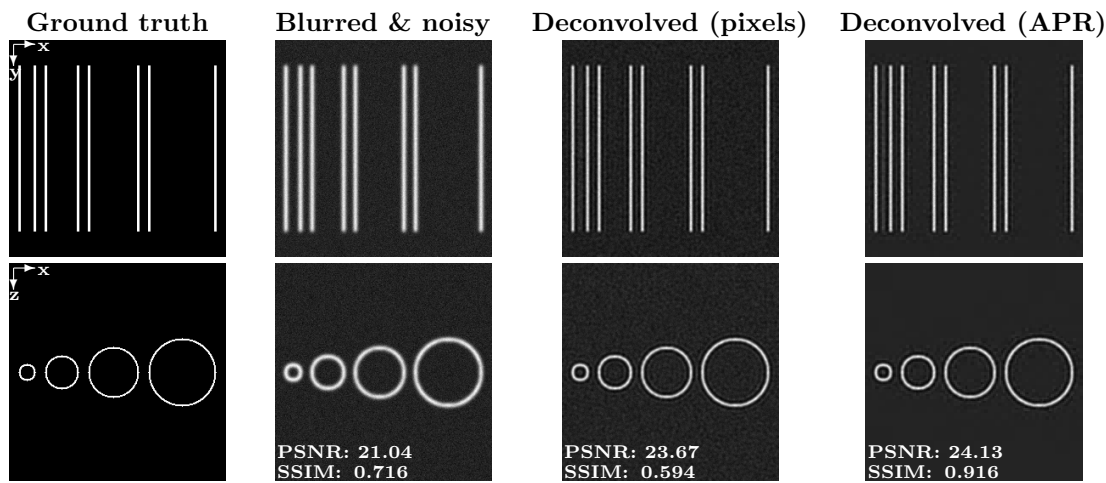

We test the resulting APR-RL deconvolution algorithm on a synthetic image of $256^3$ pixels showing hollow cylinders. For benchmark purposes, the ground-truth image is blurred using a Gaussian filter with standard deviation $\sigma=2$ pixels, truncated to a total size of $13^3$ pixels, and pixel-wise independent Gaussian noise is added. An APR is generated from the noisy and blurry benchmark image. On both test images, we perform 100 Richardson-Lucy iterations each and compare the quality of the result for pixels and APR. Figure~\ref{fig:deconvolution} shows the ground truth, blurred and noisy, and deconvolved images with RL on pixels and APR-RL. Qualitatively, the deconvolution results on pixels and on the APR look comparable. However, the pixel-based result contains pronounced noise amplification artifacts, which is typical of non-regularized RL deconvolution\cite{dey2006richardson}. This is reflected in a lower structural similarity index (SSIM) and a lower final peak signal-to-noise ratio (PSNR). The APR result avoids such artifacts in two possible ways: First, computing coarse particle values by averaging has a direct denoising effect. Second, the level-specific filter restriction tends to concentrate the filter weights around the center of the filter. Thus, convolutions applied to coarser particles increasingly resemble the identity mapping, which limits the potential for artifacts to propagate across levels. This is akin to the known error smoothing properties of multi-grid solvers in numerical analysis \cite{trottenberg2000multigrid}. 
An analysis of the convergence over RL iterations in presence and absence of noise is presented in Appendix~\ref{supp:deconv}.

The APR-RL algorithm is implemented in C++ and made available in Python via the wrapper library PyLibAPR \cite{pylibapr}. Since our GPU implementation of the APR convolution does not support stencils of size $13^3$, the experiments are run on the CPU, using the Python API. 100 RL iterations on the APR (CR=13.3) took 42 seconds with a peak memory consumption of 150\,MB. 
We implement the pixel algorithm in Python using fast Fourier transform (FFT) convolutions from the SciPy package~\cite{virtanen2020scipy}. With this implementation, 100 RL iterations on pixels required 829\,MB of memory and 180 seconds. Thus, the APR-RL algorithm is roughly 4.3 times faster and requires 5.5 times less memory. This shows that the computational performance of APR-native deconvolution can compete with an FFT-based approach on pixels, despite the high computational intensity of spatial convolution with a $13^3$ stencil.

\section{Discussion and Conclusions}

We have introduced the data structures and algorithms that enable discrete convolutions to be directly evaluated on Adaptive Particle Representations (APRs) of images using multi-core CPU or GPU acceleration. We have shown how spatial convolution can be equivalently defined on an APR by leveraging the reconstruction condition to interpolate information across resolution levels. We benchmarked the computational efficiency of our APR convolution using a set of synthetic images of varying content densities. 

For fixed image sizes, we showed that the computational performance of our CPU and GPU implementation scales with image sparsity as measured by the computational ratio (CR). APR convolution on the GPU achieved speedups of 3.8- to 28-fold compared to CUDA pixel implementations for CR values corresponding to real-world fluorescence microscopy datasets~\cite{cheeseman2018adaptive}. At the same time, APR convolutions required orders of magnitude less memory, enabling tiling-free GPU convolution of images of tens to hundreds of Gigabytes. For the largest CR$\approx$1020 tested, typical of super-resolution or expansion microscopy images, we obtained pixel-equivalent processing throughputs of up to 1\,TB/s on an inexpensive gaming GPU, which far exceeds the acquisition rate of current microscopes \cite{chhetri2015whole}. This enables complex workflows, comprising tens to hundreds of convolution operations, to be performed in real time during acquisition.

The APR convolution operation is naturally decomposed by resolution level, which offers the ability to modify the filter stencil across resolution levels. We presented two methods to do so: First, inspired by operator restriction in multi-grid methods, the filter can be downsampled to coarser resolutions, such that applying the downsampled filter on the coarser level is consistent with applying the original filter on the finest resolution. Second, the filter weights can be rescaled according to the distance between particles at each level, which is useful, e.g., in finite-difference computations. In general, the filters applied at each level can be independent of one another, providing additional design freedom for content-adaptive filters. 

Currently, our CUDA implementation is limited to filter stencils of size $3^3$ and $5^3$. The main difficulties in extending this to larger filters are the limitations to thread block size and register usage per thread. The efficiency of our implementation hinges on the ability to distribute the sparse rows corresponding to fixed positions in the $z$--$x$ plane across warps, in order to promote coalesced access to global memory. With increasing filter size, the amount of such positions quickly outgrows the maximum thread block size, such that each thread must iterate over additional rows to reconstruct isotropic patches. This adds multiples to the workload of each thread and leads to increasingly complex kernels that result in excessive register pressure, requiring compromise to avoid spillage. A possible solution is to decompose convolutions with larger stencils into sums of 1D or 2D convolutions applied line- or plane-wise~\cite{eklund2011true}.

In our benchmarks, we have neglected the time required to transfer the image data between the host computer and the GPU. Since data transfer times are proportional to the amount of data being sent, these can be expected to show similar scaling with CR as the memory usage. 
This means that in practical applications, where transfer times are significant, the overall benefit of the APR is expected to be larger than in the presented benchmarks. This is because the APR not only accelerates image processing, but also reduces the amount of data that needs to be transferred. 

Finally, we have only considered discrete convolutions. While the presented approach can in principle be extended to nonlinear filters, morphological operations, and resampling techniques, discrete convolutions remain an essential component of high-level image processing methods. For example, image features computed by convolutions are routinely used in machine learning algorithms, such as random forests and support vector machines, to perform high-level vision tasks such as object detection or instance segmentation. The ability to efficiently convolve an APR enables us to extract per-particle features and perform these tasks directly on the APR, without reverting to pixels. This not only leverages the computational ratio of the APR, but it also benefits the machine learning algorithms, as the number of learning dimensions is reduced proportionally. 

In the future, this could be exploited to design convolutional neural networks (CNNs) that directly operate on APR images using layers that implement APR convolutions. The filter weights can be learned by back-propagating the loss gradients through the interpolation across resolutions, as presented here. The filter weights for different resolution levels can be learned independently, leading to an inherent spatial scale-adaptivity of the network. The training process itself would then be guided by the spatial structure of the APR. This is in contrast to pixel CNNs, where each filter is applied uniformly across the entire image. We expect that the APR focusing on informative image regions may lead to faster convergence during training, and it may enable smaller, shallower networks. 

Taken together, APR-native discrete convolution demonstrates the potential for scalable image processing on CPUs and GPUs. The presented data structures and algorithms provide a basis for APR-native image processing and machine learning. Combined with the results from Ref.~\cite{cheeseman2018adaptive}, which showed the practical applicability of the APR to microscopy datasets, as well as existing software for generating, visualizing, and compressing APRs \cite{libapr, pylibapr}, this brings within reach complete APR-based workflows for storing, visualizing, and processing large sparse image datasets without reverting to pixels. This will allow leveraging the full potential of the APR in big bio-image projects.

\section*{Code availability} \label{sec:code}

Algorithms for the generation and manipulation of APRs are implemented in the open-source C++ software library LibAPR \cite{libapr} (available at https://github.com/mosaic-group/LibAPR). The data structures and algorithms described in this work are also implemented and publicly available there. All APR functionality is also available from Python through the wrapper library PyLibAPR \cite{pylibapr} (https://github.com/mosaic-group/PyLibAPR). The Python wrapper additionally offers interactive visualization tools for APRs (by $z$-slice and by maximum intensity projection raycast), as well as visual aids for observing the conversion of pixel images to APRs. 

\section*{Acknowledgements}
This work was supported by the Deutsche Forschungsgemeinschaft (DFG, German Research Foundation) under Germany's Excellence Strategy -- EXC-2068-390729961 -- Cluster of Excellence ``Physics of Life'' of TU Dresden, and by the Center for Scalable Data Analytics and Artificial Intelligence (ScaDS.AI) Dresden/Leipzig, funded by the Bundesministerium f\"{u}r Bildung und Forschung (BMBF, Federal Ministry of Education and Research).

\clearpage

\appendix
\section{APR convolution benchmarks supplement}

\subsection{Benchmark Images}\label{supp:images}

\begin{figure*}[htb]
    \centering
    \begin{tabular}{c c c c c}
        \textbf{CR 1.04 ($64^3$)} & \textbf{CR 3.14 ($64^3$)} & \textbf{CR 4.95 ($64^3$)} & \textbf{CR 10.77 ($128^3$)} & \textbf{CR 14.72 ($128^3$)} 
        \\
        \includegraphics[width=0.18\linewidth]{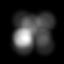}
        &
        \includegraphics[width=0.18\linewidth]{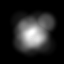}
        &
        \includegraphics[width=0.18\linewidth]{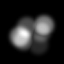}
        &
        \includegraphics[width=0.18\linewidth]{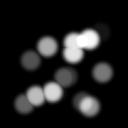}
        &
        \includegraphics[width=0.18\linewidth]{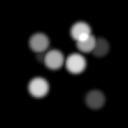}
        \\
        
        \textbf{CR 20.80 ($128^3$)} & \textbf{CR 30.27 ($128^3$)} & \textbf{CR 54.37 ($256^3$)} & \textbf{CR 124.05 ($128^3$)} & \textbf{CR 1019.83 ($256^3$)} 
        \\
        \includegraphics[width=0.18\linewidth]{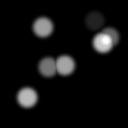}
        &
        \includegraphics[width=0.18\linewidth]{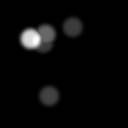}
        &
        \includegraphics[width=0.18\linewidth]{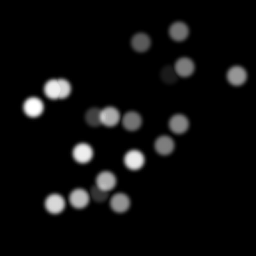}
        &
        \includegraphics[width=0.18\linewidth]{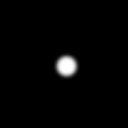}
        &
        \includegraphics[width=0.18\linewidth]{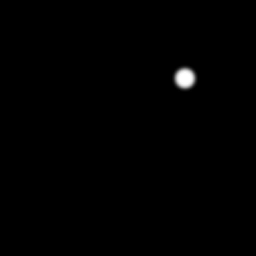}
        
    \end{tabular}
    \caption{Maximum intensity $z$-projections of the 3D benchmark images. Titles show the computational ratio (CR) and the image size in pixels in parentheses.\label{fig:benchmark_data}}
\end{figure*}

For the APR convolution benchmarks (see Section~\ref{sec:benchmarks}), we use a set of ten synthetic 3D benchmark images of varying CR as previously described~\cite{cheeseman2018adaptive}. The CR is varied by generating images that show different numbers of randomly placed spherical objects, as shown in Fig.~\ref{fig:benchmark_data} as maximum-intensity $z$-projections. The synthetic images are either $64^3$, $128^3$, or $256^3$ pixels in size. From these fixed-CR images we generate APRs of larger images by concatenating copies of the data. This generates the final benchmark images of size $512^3$ for the processing time benchmarks and of size $1024^3$ for the memory usage benchmarks. The benchmark images and their APR files are publicly available in LibAPR \cite{libapr}.

\subsection{CUDA vs. OpenMP speedup}\label{supp:benchmarks}

\begin{figure}[!ht]
    \centering
    \includegraphics[width=0.95\linewidth]{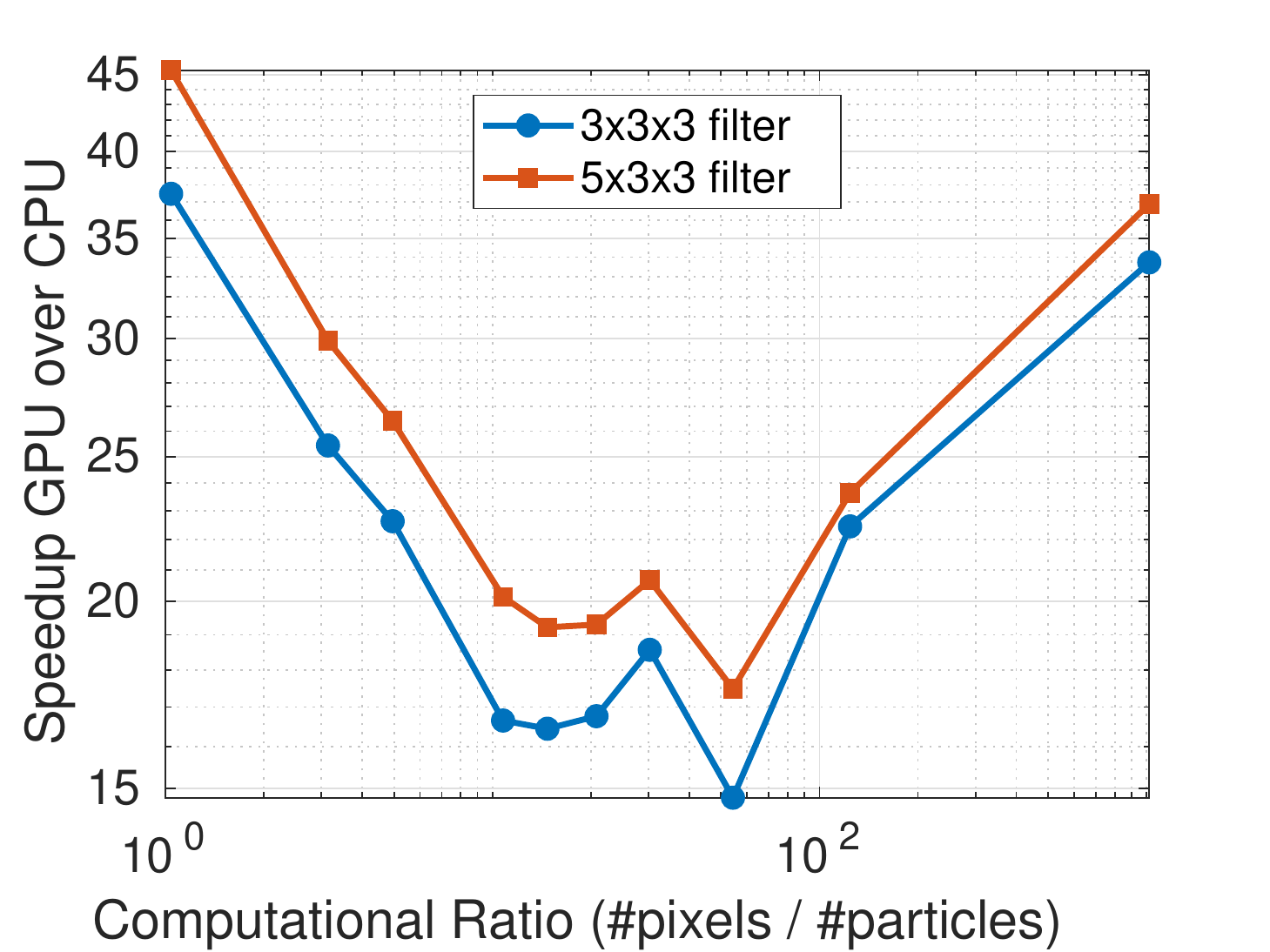}
    \caption{Speedup of APR convolution on the Nvidia GeForce RTX 2080 Max-Q GPU compared to 8-core (16 threads) parallel CPU performance on an Intel i9. Operations are timed for APRs of synthetic images at different CR values, corresponding to an image volume of $512^3$ pixels.}
    \label{fig:gpu_speedup}
\end{figure}

Figure~\ref{fig:gpu_speedup} shows the speedup of the GPU APR convolution compared to its CPU counterpart. For a CR of about 1, the GPU implementation is faster than the CPU implementation by a factor of 37 for the $3^3$ filter and 45 for the $5^3$ filter. As the CR increases past 10, the GPU speedup reduces by about a factor of two. This is likely due to increased branching (load imbalance) on the GPU. For CR values larger than 100, the GPU speedup increases again. This is where the pre-processing step pays off, as thread blocks are then only launched for non-empty rows, benefiting the GPU at high CR. 

\section{Richardson-Lucy deconvolution supplement}
\label{supp:deconv}
\begin{figure}[!ht]
    \centering
    \includegraphics[width=\linewidth]{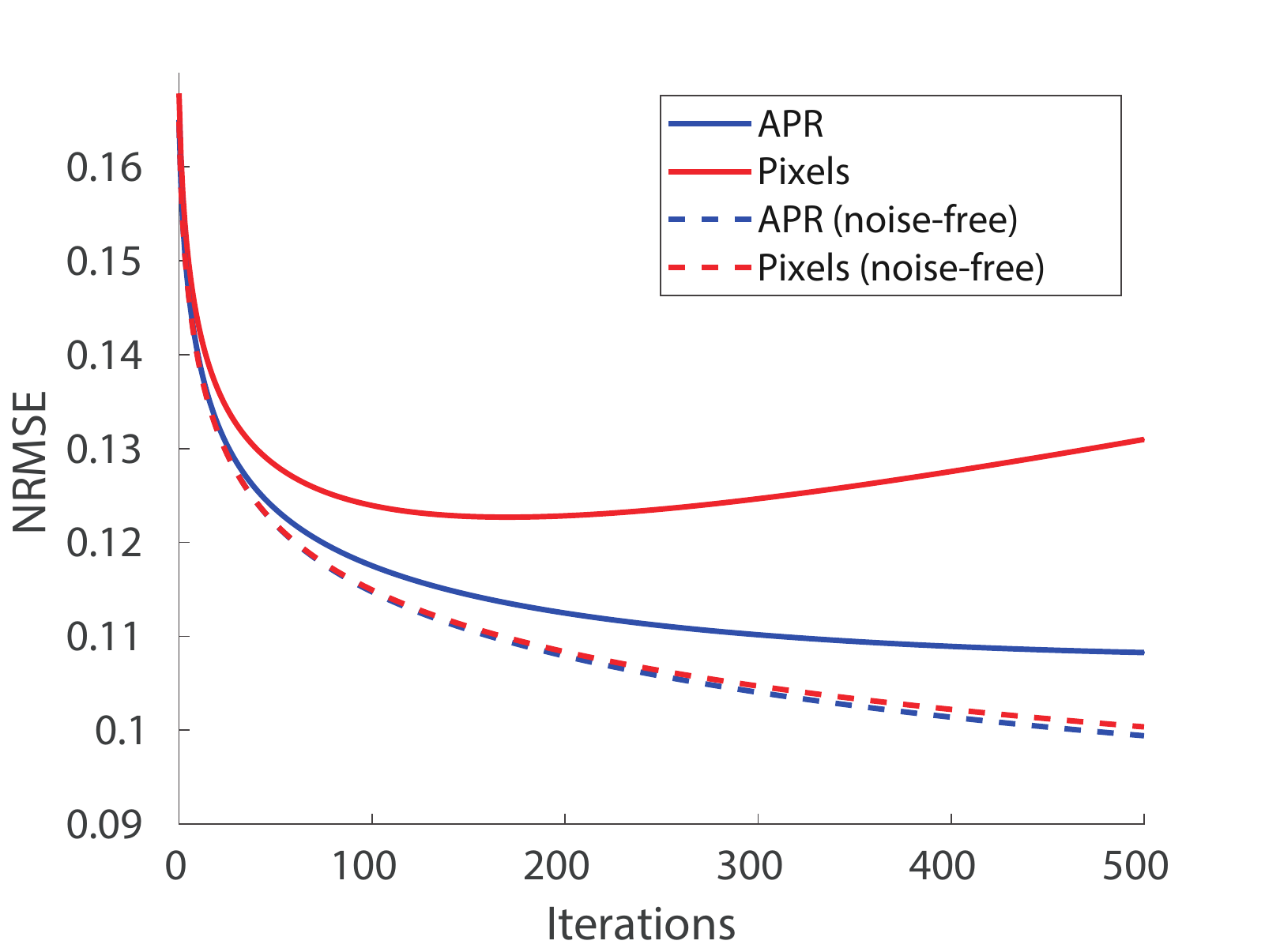}
    \caption{Normalized root mean squared error (NRMSE) of the Richardson-Lucy estimates as a function of the number of RL iterations. The results in Figure~\ref{fig:deconvolution} are after 100 iterations. Solid lines show the errors of the pixel (red) and APR (blue) algorithms on blurred and noisy input images, while the dashed lines show the corresponding errors on blurred but noise-free input images.}
    \label{fig:rl_conv}
\end{figure}

Figure~\ref{fig:rl_conv} shows the convergence of the Richardson-Lucy deconvolution example presented in Section~\ref{sec:deconv}. For the pixel algorithm applied on the blurred and noisy input image (solid red line), the error starts increasing again after about 170 iterations. This is a result of the aggregation and amplification of noise. Deconvolving the APR of the noisy input, these effects are suppressed, and the error continues to decrease throughout the 500 iterations shown here. To show that noise amplification artifacts are indeed the main contributor to this difference in performance between the APR and the pixel algorithms, we repeat the experiment using a blurred but noise-free input image. These errors are shown as dashed lines, and the pixel and APR algorithms show comparable performance and convergence. We posit that the slight edge in favor of APR-RL is due to intensity oscillation effects around the object edges.

\begin{figure}[!ht]
    \centering
    \begin{tabular}{cc}
        \textbf{Pixels (500 iterations)} & \textbf{APR (500 iterations)} \\
        \begin{overpic}[width=0.45\linewidth]{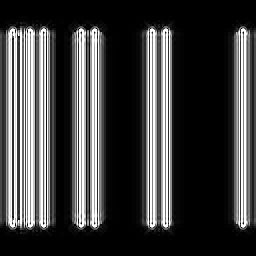}
            \put(25, 98){\color{red}\vector(0,-1){10}}	
            \put(75, 75){\color{red}\vector(-1,-2){5}}	
        \end{overpic}
        &
        \begin{overpic}[width=0.45\linewidth]{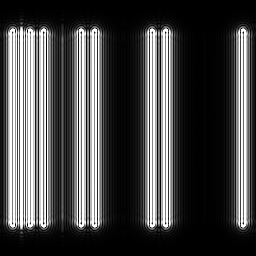}
            \put(25, 98){\color{red}\vector(0,-1){10}}	
            \put(75, 75){\color{red}\vector(-1,-2){5}}	
        \end{overpic}
    \end{tabular}
    \caption{High-contrast images of the pixel-wise relative error after 500 iterations of Richardson-Lucy deconvolution on pixels (left) and on APR (right). The intensity oscillation (ringing) artifacts (pointed out by red arrows) are suppressed in coarsely sampled regions of the APR, possibly explaining the slightly lower error for noise-free input images in Figure~\ref{fig:rl_conv}.}
    \label{fig:rl_error}
\end{figure}

Indeed, the pixel-based algorithm on this test image creates intensity oscillation (ringing) artifacts around the bright edges in the image. These artifacts then spread outward from the edges and become amplified as the iterations progress. For the APR-based deconvolution algorithm, this effect is suppressed in the coarsely sampled image regions. Figure~\ref{fig:rl_error} shows contrast-enhanced images of the relative error after 500 iterations of both methods to illustrate this.

\begin{thebibliography}{10}

\bibitem{reynaud2014guide}
Emmanuel~G Reynaud, Jan Peychl, Jan Huisken, and Pavel Tomancak.
\newblock Guide to light-sheet microscopy for adventurous biologists.
\newblock {\em Nature methods}, 12(1):30, 2014.

\bibitem{mcdole2018toto}
Katie McDole, L{\'e}o Guignard, Fernando Amat, Andrew Berger, Gr{\'e}goire
  Malandain, Lo\"{\i}c~A Royer, Srinivas~C Turaga, Kristin Branson, and
  Philipp~J Keller.
\newblock In toto imaging and reconstruction of post-implantation mouse
  development at the single-cell level.
\newblock {\em Cell}, 175(3):859--876, 2018.

\bibitem{huisken2012slicing}
Jan Huisken.
\newblock Slicing embryos gently with laser light sheets.
\newblock {\em Bioessays}, 34(5):406--411, 2012.

\bibitem{chhetri2015whole}
Raghav~K Chhetri, Fernando Amat, Yinan Wan, Burkhard H{\"o}ckendorf, William~C
  Lemon, and Philipp~J Keller.
\newblock Whole-animal functional and developmental imaging with isotropic
  spatial resolution.
\newblock {\em Nature methods}, 12(12):1171--1178, 2015.

\bibitem{allan2012omero}
Chris Allan, Jean-Marie Burel, Josh Moore, Colin Blackburn, Melissa Linkert,
  Scott Loynton, Donald MacDonald, William~J Moore, Carlos Neves, Andrew
  Patterson, et~al.
\newblock {OMERO}: flexible, model-driven data management for experimental
  biology.
\newblock {\em Nature methods}, 9(3):245--253, 2012.

\bibitem{gunther2019scenery}
U.~{G{\"u}nther}, T.~{Pietzsch}, A.~{Gupta}, K.~I.~S. {Harrington},
  P.~{Tomancak}, S.~{Gumhold}, and I.~F. {Sbalzarini}.
\newblock Scenery: flexible virtual reality visualization on the {J}ava {VM}.
\newblock In {\em 2019 IEEE Visualization Conference (VIS)}, pages 1--5, 2019.

\bibitem{schindelin2012fiji}
Johannes Schindelin, Ignacio Arganda-Carreras, Erwin Frise, Verena Kaynig, Mark
  Longair, Tobias Pietzsch, Stephan Preibisch, Curtis Rueden, Stephan Saalfeld,
  Benjamin Schmid, et~al.
\newblock Fiji: an open-source platform for biological-image analysis.
\newblock {\em Nature methods}, 9(7):676--682, 2012.

\bibitem{afshar2016parallel}
Yaser Afshar and Ivo~F Sbalzarini.
\newblock A parallel distributed-memory particle method enables
  acquisition-rate segmentation of large fluorescence microscopy images.
\newblock {\em PloS one}, 11(4):e0152528, 2016.

\bibitem{royer2015clearvolume}
Loic~A Royer, Martin Weigert, Ulrik G{\"u}nther, Nicola Maghelli, Florian Jug,
  Ivo~F Sbalzarini, and Eugene~W Myers.
\newblock Clear{V}olume: open-source live 3{D} visualization for light-sheet
  microscopy.
\newblock {\em Nature methods}, 12(6):480--481, 2015.

\bibitem{haase2020clij}
Robert Haase, Loic~A Royer, Peter Steinbach, Deborah Schmidt, Alexandr Dibrov,
  Uwe Schmidt, Martin Weigert, Nicola Maghelli, Pavel Tomancak, Florian Jug,
  et~al.
\newblock {CLIJ}: {GPU}-accelerated image processing for everyone.
\newblock {\em Nature methods}, 17(1):5--6, 2020.

\bibitem{moore2021ome}
Josh Moore, Chris Allan, Sebastien Besson, Jean-Marie Burel, Erin Diel, David
  Gault, Kevin Kozlowski, Dominik Lindner, Melissa Linkert, Trevor Manz, et~al.
\newblock {OME-NGFF}: scalable format strategies for interoperable bioimaging
  data.
\newblock {\em BioRxiv}, 2021.

\bibitem{pietzsch2015bigdataviewer}
Tobias Pietzsch, Stephan Saalfeld, Stephan Preibisch, and Pavel Tomancak.
\newblock {B}ig{D}ata{V}iewer: visualization and processing for large image
  data sets.
\newblock {\em Nature methods}, 12(6):481--483, 2015.

\bibitem{bria2016terafly}
Alessandro Bria, Giulio Iannello, Leonardo Onofri, and Hanchuan Peng.
\newblock {T}era{F}ly: real-time three-dimensional visualization and annotation
  of terabytes of multidimensional volumetric images.
\newblock {\em Nature methods}, 13(3):192--194, 2016.

\bibitem{amat2015efficient}
Fernando Amat, Burkhard H{\"o}ckendorf, Yinan Wan, William~C Lemon, Katie
  McDole, and Philipp~J Keller.
\newblock Efficient processing and analysis of large-scale light-sheet
  microscopy data.
\newblock {\em Nature protocols}, 10(11):1679, 2015.

\bibitem{balazs2017real}
B{\'a}lint Bal{\'a}zs, Joran Deschamps, Marvin Albert, Jonas Ries, and Lars
  Hufnagel.
\newblock A real-time compression library for microscopy images.
\newblock {\em bioRxiv}, page 164624, 2017.

\bibitem{cheeseman2018adaptive}
Bevan~L Cheeseman, Ulrik G{\"u}nther, Krzysztof Gonciarz, Mateusz Susik, and
  Ivo~F Sbalzarini.
\newblock Adaptive particle representation of fluorescence microscopy images.
\newblock {\em Nature communications}, 9(1):5160, 2018.

\bibitem{rivron2018blastocyst}
Nicolas~C Rivron, Javier Frias-Aldeguer, Erik~J Vrij, Jean-Charles Boisset,
  Jeroen Korving, Judith Vivi{\'e}, Roman~K Truckenm{\"u}ller, Alexander {Van
  Oudenaarden}, Clemens~A {Van Blitterswijk}, and Niels Geijsen.
\newblock Blastocyst-like structures generated solely from stem cells.
\newblock {\em Nature}, 557(7703):106--111, 2018.

\bibitem{ljosa2012annotated}
Vebjorn Ljosa, Katherine~L Sokolnicki, and Anne~E Carpenter.
\newblock Annotated high-throughput microscopy image sets for validation.
\newblock {\em Nature methods}, 9(7):637--637, 2012.

\bibitem{adelson1984pyramid}
Edward~H Adelson, Charles~H Anderson, James~R Bergen, Peter~J Burt, and Joan~M
  Ogden.
\newblock Pyramid methods in image processing.
\newblock {\em RCA engineer}, 29(6):33--41, 1984.

\bibitem{meagher1982geometric}
Donald Meagher.
\newblock Geometric modeling using octree encoding.
\newblock {\em Computer graphics and image processing}, 19(2):129--147, 1982.

\bibitem{porwik2004haar}
Piotr Porwik and Agnieszka Lisowska.
\newblock The {H}aar-wavelet transform in digital image processing: its status
  and achievements.
\newblock {\em Machine graphics and vision}, 13(1/2):79--98, 2004.

\bibitem{libapr}
Bevan Cheeseman, Krzysztof Gonciarz, Ulrik G{\"u}nther, Joel Jonsson, Mario
  Emmenlauer, and Mateusz Susik.
\newblock cheesema/libapr: Initial release v1.1, September 2018.

\bibitem{haber2018learning}
Eldad Haber, Lars Ruthotto, Elliot Holtham, and Seong-Hwan Jun.
\newblock Learning {A}cross {S}cales---{M}ultiscale {M}ethods for {C}onvolution
  {N}eural {N}etworks.
\newblock In {\em Proceedings of the AAAI Conference on Artificial
  Intelligence}, volume~32, 2018.

\bibitem{trottenberg2000multigrid}
Ulrich Trottenberg, Cornelius~W Oosterlee, and Anton Schuller.
\newblock {\em Multigrid}.
\newblock Elsevier, 2000.

\bibitem{zhang2018poisson}
Yide Zhang, Yinhao Zhu, Evan Nichols, Qingfei Wang, Siyuan Zhang, Cody Smith,
  and Scott Howard.
\newblock A {P}oisson-{G}aussian denoising dataset with real fluorescence
  microscopy images.
\newblock {\em arXiv preprint arXiv:1812.10366}, 2018.

\bibitem{dagum1998openmp}
Leonardo Dagum and Ramesh Menon.
\newblock {OpenMP}: an industry standard {API} for shared-memory programming.
\newblock {\em IEEE computational science and engineering}, 5(1):46--55, 1998.

\bibitem{Yalamanchili2015}
Pavan Yalamanchili, Umar Arshad, Zakiuddin Mohammed, Pradeep Garigipati, Peter
  Entschev, Brian Kloppenborg, James Malcolm, and John Melonakos.
\newblock {ArrayFire - A high performance software library for parallel
  computing with an easy-to-use API}, 2015.

\bibitem{richardson1972bayesian}
William~Hadley Richardson.
\newblock Bayesian-based iterative method of image restoration.
\newblock {\em JoSA}, 62(1):55--59, 1972.

\bibitem{lucy1974iterative}
Leon~B Lucy.
\newblock An iterative technique for the rectification of observed
  distributions.
\newblock {\em The astronomical journal}, 79:745, 1974.

\bibitem{dey2006richardson}
Nicolas Dey, Laure Blanc-Feraud, Christophe Zimmer, Pascal Roux, Zvi Kam,
  Jean-Christophe Olivo-Marin, and Josiane Zerubia.
\newblock Richardson--{L}ucy algorithm with total variation regularization for
  3{D} confocal microscope deconvolution.
\newblock {\em Microscopy research and technique}, 69(4):260--266, 2006.

\bibitem{pylibapr}
Joel Jonsson, Bevan Cheeseman, and Krzysztof Gonciarz.
\newblock mosaic-group/pylibapr, 2021.

\bibitem{virtanen2020scipy}
Pauli Virtanen, Ralf Gommers, Travis~E Oliphant, Matt Haberland, Tyler Reddy,
  David Cournapeau, Evgeni Burovski, Pearu Peterson, Warren Weckesser, Jonathan
  Bright, et~al.
\newblock Sci{P}y 1.0: fundamental algorithms for scientific computing in
  {P}ython.
\newblock {\em Nature methods}, 17(3):261--272, 2020.

\bibitem{eklund2011true}
Anders Eklund, Mats Andersson, and Hans Knutsson.
\newblock True 4{D} image denoising on the {GPU}.
\newblock {\em International Journal of Biomedical Imaging}, 2011, 2011.

\end{thebibliography}
\end{document}